# Neural Network Assisted Lifting Steps For Improved Fully Scalable Lossy Image Compression in JPEG 2000

Xinyue Li, *Student Member, IEEE,* Aous Naman, *Senior Member, IEEE,* and David Taubman, *Fellow, IEEE*

**Abstract**—This work proposes to augment the lifting steps of the conventional wavelet transform with additional neural network assisted lifting steps. These additional steps reduce residual redundancy (notably aliasing information) amongst the wavelet subbands, and also improve the visual quality of reconstructed images at reduced resolutions. The proposed approach involves two steps, a high-to-low step followed by a low-to-high step. The high-to-low step suppresses aliasing in the low-pass band by using the detail bands at the same resolution, while the low-to-high step aims to further remove redundancy from detail bands, so as to achieve higher energy compaction. The proposed two lifting steps are trained in an end-to-end fashion; we employ a backward annealing approach to overcome the non-differentiability of the quantization and cost functions during back-propagation. Importantly, the networks employed in this paper are compact and with limited non-linearities, allowing a fully scalable system; one pair of trained network parameters are applied for all levels of decomposition and for all bit-rates of interest. By employing the proposed approach within the JPEG 2000 image coding standard, our method can achieve up to 17.4% average BD bit-rate saving over a wide range of bit-rates, while retaining quality and resolution scalability features of JPEG 2000.

## 1 Introduction

The wavelet transform has been successfully employed in a variety of codecs and open image compression standards; examples include JPEG 2000 [1] [2], the BBC's VC2 codec, and JPEG-XS [3]. The wavelet transform provides a balance between energy compaction and sparsity preservation, by analyzing the image with a hierarchical family of compact support operators, realized through successive filtering and down-sampling. Importantly, the wavelet transform naturally produces a multi-resolution representation of the image, which enables reconstructions at dyadically-spaced image resolutions, a feature known as resolution scalability.

Although the wavelet transform provides excellent energy compaction for horizontal and vertical edges, slanted features are poorly characterized by the separable wavelet filters, which leads to significant redundancy between all subbands and visually disturbing artifacts in the reconstructed images along diagonal edges. Solutions have been explored to improve directional sensitivity of the wavelet transform, which can be broadly categorized into traditional approaches and machine-learning based methods.

In the traditional approaches, oriented wavelet transforms employing directional filter banks are proposed to capture geometric structures within an image; examples include the Contourlet Transform [4], the Curvelet Transform [5], the Curved Wavelet Transform [6], the Ridgelet Transform [7], the Bandlet Transform [8], and directional wavelet transforms [9] [10]. However, when such schemes are employed for image compression, the orientation information needs to be explicitly coded and communicated in order to correctly inverse the respective transform.

In the last decade, researchers experimented with machine learning (ML) based approaches to improve coding efficiency in image and video compression applications, with very promising results. For the wavelet-based image compression, which is the topic of this work, examples include [11], [12], [13], [14], [15]. In [11], Ma et al. propose a neural network for context modeling in their JPEG2000-inspired arithmetic coder, which they identify as Pixel Convolutional Neural Network (PixelCNN); they also propose a post-processing step to enhance reconstructed image quality. In a later work [12], they propose an *iWave* transform; this transform replaces the predict step of the conventional wavelet transform with a CNN while keeping the update step as a simple averaging operation. The iWave transform improves energy compaction compared to the CDF 9/7-based wavelet transform of the JPEG2000 standard. In [13], Dardouri et al. propose to replace both the predict and update steps of the conventional wavelet transform with a Fully Connected Neural Network. This work is further extended in [14]; however, performance improvements over JPEG2000 could be obtained only for the SSIM metric and the uncommon PieAPP metric. In [15], Li et al. propose the reversible autoencoder (Rev-AE), which is a lifting based wavelet-like codec with theoretical guarantees on transform reversibility and robustness to reconstruction quantization errors; the proposed approach shows competitive results compared to JPEG2000.

These methods inherit the multi-scale representation from the wavelet transform, which provides resolution scalability; however, none of them explore quality scalability or region-of-interest accessibility. Additionally these works do not investigate ways to directly train the networks for rate-distortion; instead, alternative training objectives, such as energy compaction of the transformed coefficients or pre-diction residuals, are used as proxies for coding efficiency.

In contrary, some researchers experimented with image compression designs that employ neural networks only; these designs usually adopt an end-to-end optimization that explicitly targets rate-distortion objectives. Ballè et al. [16] introduce a general compression framework for end-to-end rate–distortion optimization, in which generalized divisive normalization [17] is employed to reduce mutual information between transformed channels. Subsequently, they introduce a scale hyper-prior in [18] to effectively capture spatial dependencies in the latent representation; similar approaches can also be found in [19]. To improve this CNN based auto-encoder, Theis et al [20] propose to employ Gaussian scale mixtures to model the probability density function of coded coefficients and estimate their entropy; the entropy is then used to estimates bit-rates and drive the backpropagation-based training. Meanwhile, Toderici et al. first introduce the Recurrent Neural Network (RNN) in an end-to-end optimization scheme [21] [22]. This work is further improved by Johnsons et al. using spatially adaptive bit allocation and SSIM weighted loss function [23]. Other works can be found in [24], [25], [26], [27], [28].

Even though these end-to-end schemes achieve significantly better compression results, they lack resolution scalability, quality scalability and region-of-interest accessibility of wavelet-based compression frameworks. They also have significantly higher computational complexity and huge receptive field in the image domain. Additionally, network structures and trained parameters are mostly dependent on the target compression bit-rates.

The goal of our work is to develop a low-complexity neural-network-assisted compression scheme, which inherits all the important attributes from the conventional wavelet-based frameworks, while also taking the advantage of end-to-end rate-distortion optimization. Specifically, we propose two neural network assisted lifting steps in addition to the lifting steps of the conventional wavelet transform to exploit residual redundancy amongst the wavelet subbands. The first step, identified here as a *high-to-low* step, aims to estimate and remove redundancy (notably aliasing information) in the low-pass band produced at each level of the transform, utilizing the detail bands at the same scale. This step also improves the visual quality for the modified LL band at each level of decomposition. The second step, identified here as a *low-to-high* step, aims to further compact the high frequency coefficients of the wavelet transform, so as to reduce redundancy among the detail subbands.

In addition, we propose a comprehensive training strategy to jointly train these two lifting steps in an end-to-end fashion, leading to higher coding efficiency. These trained networks are applied uniformly to all levels in the wavelet decomposition and to all the compression bit-rates of interest. This means that our method is fully scalable, with no need to learn and store separate network parameters for each decomposition level or for different bit-rates.

Rather than simply presenting the proposed methodology, Section 2 of this paper studies the underlying problem associated with removing redundancy from the low- and high-pass subbands. This study explicitly shows that geometric flow provides opportunities to untangle aliasing and other sources of redundancy, using a bank of linear operators controlled dynamically by opacities. This analysis underpins the proposed approach, and interestingly we find that the compression performance of the various investigated neural network structures is enhanced if we follow the suggestions of the underlying theory; this is the main contribution of Section 4.

The remainder of this paper is organized as follows. Section 3 summarizes three generic architectures of the proposed method. Section 5 elaborates the comprehensive training strategy that we propose to jointly optimize the networks in an end-to-end fashion. Experimental results and comparisons with existing methods are shown in Section 6, followed by conclusions and discussions in Section 7.

## 2 Motivation and Fundamentals

This section highlights the generic problems and opportunities associated with reducing residual redundancy in the wavelet transform, especially with the aid of geometric flow in the two-dimensional (2D) scenario.

Although deterministic redundancy, i.e. oversampling, is avoided in the wavelet representation, statistical redundancy, especially the aliasing-related residual redundancy, is still inevitably present amongst the wavelet subbands. This is because the wavelet transform imposes strong conditions on the critically sampled filter banks, which prevents the redundancy from being eliminated between different subbands. Specifically, the analysis and synthesis filters $h_0$ and $g_0$ of a two-channel critically sampled filter bank must satisfy the following constraint in the Fourier domain:

$$\hat{h}_0(\omega)\hat{g}_0(\omega) + \hat{h}_0(\pi - \omega)\hat{g}_0(\pi - \omega) = 1 \tag{1}$$

which means in particular that $\hat{h}_0(\pi - \omega)\hat{g}_0(\pi - \omega) = \frac{1}{2}$ at $\omega = \frac{\pi}{2}$. Since finite support filters must have continuous transfer functions, the low-pass analysis filter $h_0$ must have a significant response to frequencies $\omega > \frac{\pi}{2}$, which corresponds to aliasing in the low-pass subband. This aliasing content is both visually disturbing and a form of redundancy. Similarly the high-pass analysis filter $h_1$, which is in mirror symmetry with $g_0$, necessarily has a significant response to frequencies $\omega < \frac{\pi}{2}$. This pollution of the high-pass subband with low frequency content is another form of information redundancy between the two subbands.

The goal of this paper is to reduce this redundant content between the low- and high-pass subbands in the wavelet transform. There are many possible ways to achieve this. One can attempt to improve the subband filters themselves, but no amount of improvement in the filters can escape from the implications of (1), which indicates that $h_0$ cannot be designed to avoid aliasing. If $h_0$ does manage to roll off close to 0 by $\frac{\pi}{2}$, then $g_0$ would require a huge gain around the half-band frequency $\frac{\pi}{2}$, greatly amplifying quantization errors. The same is true for the high-pass analysis and synthesis filters $h_1$ and $g_1$. Although designs involving longer filters can have smaller transition bands around $\frac{\pi}{2}$, this comes at the cost of a loss of sparsity – innovative features such as edges in the space domain produce more non-zero subband samples, which adversely impact coding efficiency.

More generally, additional operators could be introduced to untangle the redundant information amongst the wavelet coefficients, which is the main focus of this paper.

Let $\mathcal{A}_L(x)$ and $\mathcal{A}_H(x)$ denote the analysis of signal $x$ into the low-pass band $y_L = \mathcal{A}_L(x)$ and the high-pass (detail) subband $y_H = \mathcal{A}_H(x)$ respectively, within one level of a Discrete Wavelet Transform (DWT). For a 2D DWT following the Mallat decomposition structure, $y_H$ stands for the collection of all three detail subbands, denoted as HL, LH and HH, but most of the material in this section is most easily presented in 1D in the first instance.

In particular, suppose an operator $\mathcal{T}^A_{H2L}$ can be found to estimate the aliased component $\tilde{y}_L$ of $y_L$ using $y_H$, written as $\tilde{y}_L = \mathcal{T}^A_{H2L}(y_H)$, then the non-aliased component $\bar{y}_L$ can be separated from $y_L$ as $\bar{y}_L = y_L - \tilde{y}_L$. Since $\bar{y}_L$ is at least approximately free from aliasing, now all of the aliasing information $\tilde{y}_H$ inside $y_H$ arises from the content in $\bar{y}_L$. This means that $\bar{y}_L$ can then be used to discover the aliasing contribution within $y_H$, written as $\tilde{y}_H = \mathcal{T}^W_{L2H}(\bar{y}_L)$. In fact, the operator $\mathcal{T}^W_{L2H}$ can simply be an LSI filter, because $\tilde{y}_H$ should ideally be equal to $\mathcal{A}_H(\mathcal{I}(\bar{y}_L))$, where $\mathcal{I}$ stands for the ideal interpolator. We have chosen to use the superscript $^W$ for this second operator, to emphasize the fact that it could potentially be obtained as a conventional Wiener filter.

In this scenario, the development of the operator $\mathcal{T}^A_{H2L}$ is the one that presents the greatest challenge; it cannot simply be an LSI filter. If it were, then the transformation steps $\bar{y}_L = y_L - \mathcal{T}^A_{H2L}(y_H)$, followed by $\bar{y}_H = y_H - \mathcal{T}^W_{L2H}(\bar{y}_L)$ could be seen as augmenting the original wavelet transform with two additional LSI lifting steps, so that the complete transform is equivalent to choosing a different set of subband analysis and synthesis filters, which we have already ruled out as a viable solution. As we shall see, the $\mathcal{T}^A_{H2L}$ should at least be adaptive to local geometric structure, and we use the superscript $^A$ here to highlight both its role in untangling aliasing and the need for local adaptivity.

From a different perspective, suppose an operator $\mathcal{T}^A_{L2H}$ can be developed to discover the aliased part $\tilde{y}_H$ of $y_H$ using $y_L$, written as $\tilde{y}_H = \mathcal{T}^A_{L2H}(y_L)$. Then the "cleaned" high-pass band $\bar{y}_H = y_H - \tilde{y}_H$ can be used to untangle the aliasing information $\tilde{y}_L$ in $y_L$ using an operator $\mathcal{T}^W_{H2L}$; that is $\tilde{y}_L = \mathcal{T}^W_{H2L}(\bar{y}_H)$. In this converse scenario, the operator $\mathcal{T}^W_{H2L}$ becomes conceptually simple, potentially being a Wiener filter, whereas $\mathcal{T}^A_{L2H}$ cannot be LSI.

The two perspectives demonstrate that at least one of these two operators $\mathcal{T}^A_{L2H}$ and $\mathcal{T}^A_{H2L}$ is difficult to develop, depending on which we choose to perform first. Although these two approaches may seem equally plausible considering only one level of DWT decomposition, the difference appears with multiple levels of decomposition. This will be further discussed in Section 3.

For the moment, considering only one level of decomposition, we first examine the fundamental difficulties and opportunities to untangle the aliasing in the high-pass band using the low-pass band, i.e. the operator $\mathcal{T}^A_{L2H}$. Although there is no general deterministic way to construct $\mathcal{T}^A_{L2H}$ to untangle the aliasing, prior statistical signal models can be used to derive a posterior distribution for the aliasing component, from which an estimate can be formed. This is essentially the basis of super-resolution algorithms, for which the key challenge is to estimate original high frequency components that appear as aliasing in a low resolution source image.

Estimating the aliasing component of a signal is much easier to do in the image domain than in one dimension, since geometric flow in images provides a strong form of prior knowledge. Specifically, we expect that edges in the underlying spatially continuous image are smooth along their contours, so that the innovative aspects of an edge, namely its profile, change only slowly along the edge, i.e. along the geometric flow. This geometric regularity provides an opportunity to untangle aliasing in the 2D DWT.

We can see this more concretely by considering a continuous and consistently oriented signal $f(s_1, s_2)$, such that the edge profile is exactly the same along an orientation with slope $\alpha$ as shown in Fig. 1. This 2D continuous signal $f(s_1, s_2)$ can be understood as an ensemble of multiple shifted copies of the prototype 1D signal $f(s_1)$. That is $f(s_1, s_2) \equiv f_{s_2}(s_1) = f(s_1 - \alpha \cdot s_2)$, where $f_{s_2}$ represents the horizontal cross-section of $f$ at the vertical position $s_2$ as highlighted in Fig. 1. In our scenario, the 2D underlying continuous signal $f$ is a Nyquist band-limited image, whose samples correspond to the discrete image $x$. To model the discrete wavelet transformation of $x$, $f$ is then subjected to the continuous analogues of the wavelet analysis low-pass filter $h_L$ and high-pass filter $h_H$, producing low- and high-pass images $f_L$ and $f_H$ respectively.

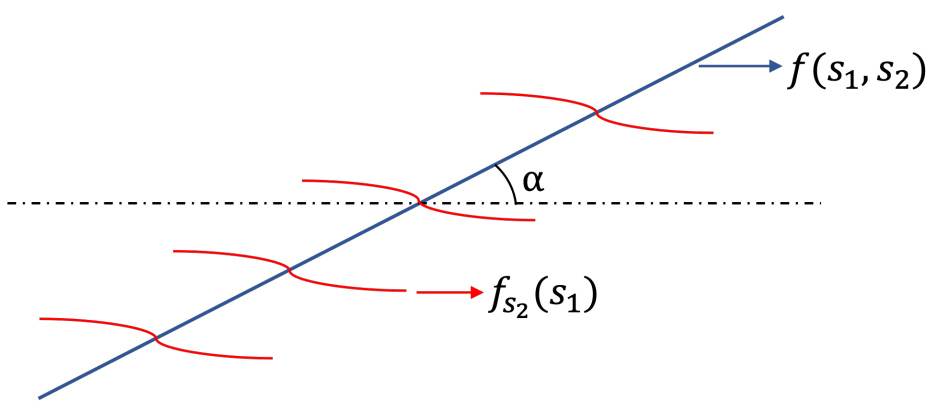

Fig. 1. The illustration of an orientated image feature along with its geometric flow, as highlighted in red.

The cross-section $f_{L,s_2}$ of $f_L$ and its discrete counterpart $x_{L,n_2}$ can be written in the horizontal Fourier domain as

$$\begin{aligned}\hat{f}_{L,s_2}(\omega) &= \hat{h}_L(\omega)\hat{f}(\omega)e^{-j\alpha s_2 \omega} \\ &= \hat{h}_L(\omega)\hat{x}(\omega)e^{-j\alpha n_2 \omega} = \hat{x}_{L,n_2}(\omega)\end{aligned}$$

The discrete wavelet low-pass subband $y_{L,n_2}$ is just a subsampled version of $x_{L,n_2}$; considering only one level of decomposition, $y_{L,n_2}$ can be written as

$$\begin{aligned}\hat{y}_{L,n_2}(\omega) &= \tfrac{1}{2}\hat{x}_{L,n_2}(\tfrac{\omega}{2}) + \tfrac{1}{2}\hat{x}_{L,n_2}(\tfrac{\omega}{2} - \pi) \\ &= \tfrac{1}{2}\hat{h}_L(\tfrac{\omega}{2})\hat{x}(\tfrac{\omega}{2})e^{-j\alpha n_2 \omega/2} \\ &\quad + \tfrac{1}{2}\hat{h}_L(\tfrac{\omega}{2} - \pi)\hat{x}(\tfrac{\omega}{2} - \pi)e^{-j\alpha n_2(\omega/2 - \pi)}\end{aligned}$$

which reveals its aliased and non-aliased components.

Averaging the inverse shifted signals over a vertical neighborhood $\mathcal{N}_2$ yields

$$\begin{aligned}\bar{y}_L(\omega) &= \frac{2}{\|\mathcal{N}_2\|}\sum_{\mathcal{N}_2}\hat{y}_{L,n_2}(\omega)e^{j\alpha n_2 \omega/2} \\ &= \hat{h}_L(\tfrac{\omega}{2})\hat{x}(\tfrac{\omega}{2}) + \frac{2}{\|\mathcal{N}_2\|}\sum_{\mathcal{N}_2}\hat{h}_L(\tfrac{\omega}{2} - \pi)\hat{x}(\tfrac{\omega}{2} - \pi)e^{j\alpha n_2 \pi}\end{aligned}$$

The last term above averages aliasing components and can be expected to be small, so long as $\alpha$ is not an integer and the averaging neighbourhood is sufficiently large. As a result, $\bar{y}_L(\omega) \approx \hat{h}_L(\omega/2)\hat{x}(\omega/2)$. Once aliasing components are effectively untangled, $\bar{y}_L$ can then be used to estimate the aliasing contribution $\tilde{y}_H$ in the high-pass subband $y_H$ using an LSI filter. This entire process, starting from $y_L$ to untangle $\tilde{y}_H$, provides a viable solution for constructing $\mathcal{T}^A_{L2H}$.

Moreover, $\bar{y}_L$ can be combined with $y_L$ to recover an estimate of the original image $f_L$. This demonstrates the connection between untangling aliasing from a low-pass subband and the well studied problem of super resolution. More generally, the simple averaging process suggested above can be replaced by a Wiener filter. As shown in [29], given multiple aliased views of the same underlying continuous image, where each view is obtained with a different shift, the minimum mean squared error best estimate of the original scene can indeed be found using Wiener filtering.

That is, the problem of untangling aliasing can in fact be solved using a filter-based strategy, so long as we can identify multiple copies of the same underlying feature, with known shifts between each copy – i.e. known geometric flow. Since geometric flow is a local property within an image, the untangling of aliasing requires either an adaptive filtering solution or a bank of filters with an adaptive strategy for combining their responses, so the overall operator $\mathcal{T}^A_{L2H}$ cannot be LSI and will generally need to be non-linear. As we shall see in Section 4, this is essentially the structure that we have found to work best. Although this discussion has been limited to the case in which we start from the low-pass subband[1], the dual problem, in which the first step uses the high-pass subband to discover and clean the redundant aliasing information, has exactly the same properties.

Now the challenges are: 1) how to discover geometric flow from the aliased content in the subband domain, while determining whether or not usable structure is actually present; and 2) making the steps in the transform as robust to quantization noise as possible. The first challenge is addressed by the adoption of a suitable network structure, as developed in Section 4. The second challenge arises because the transform and its inverse are to be used in a compression system; noting that the steps that untangle aliasing are necessarily non-linear, quantization errors are magnified in a data-dependent way. This is addressed by a disciplined learning strategy detailed in Section 5.

Before coming to these challenges, Section 3 explores the construction of invertible transforms based on the operators $\mathcal{T}_{H2L}$ and $\mathcal{T}_{L2H}$, specifically addressing the implications of starting with $\mathcal{T}^A_{H2L}$ (high-to-low approach) versus $\mathcal{T}^A_{L2H}$ (low-to-high approach).

## 3 Transform Structures

This section summarizes three generic architectures which can exploit geometric flow and untangle aliasing content within the wavelet subbands; we refer to these as low-to-high, high-to-low and hybrid approaches.

The low-to-high approach aims to suppress redundant information within the detail bands HL, LH and HH with the aid of the low-pass (LL) band from the same decomposition level, as illustrated in Fig. 2. We do this using an operator $\mathcal{T}^A_{L2H}$, which can be understood as forming a prediction of HL, LH and HH from the LL band. More specifically, we expect this operator to be able to exploit local geometric flow to predict the aliased components within HL, LH and HH, as explained above. Conceptually, if the operator $\mathcal{T}^A_{L2H}$ completely removes redundancy within the detail bands, then further cleaning aliasing $\tilde{\mathbf{y}}_{LL}$ in the LL band can be achieved simply using a linear operator $\mathcal{T}^W_{H2L}$ as explained in Section 2.

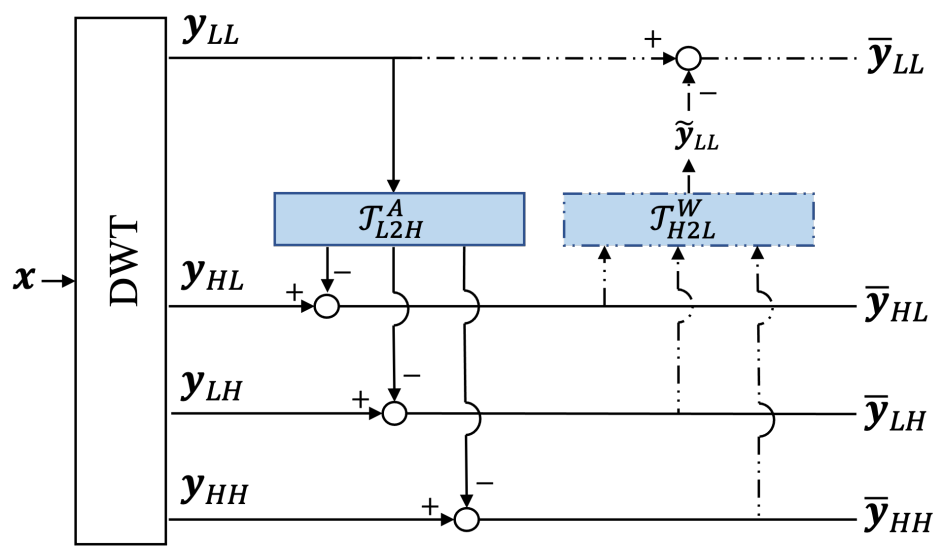

Fig. 2. The architecture of the low-to-high approach. The symbols $\mathbf{y}_{LL}$, $\mathbf{y}_{HL}$, $\mathbf{y}_{LH}$ and $\mathbf{y}_{HH}$ represent the LL, HL, LH and HH bands of the wavelet transform. The symbols $\bar{\mathbf{y}}_{LL}$, $\bar{\mathbf{y}}_{HL}$, $\bar{\mathbf{y}}_{LH}$ and $\bar{\mathbf{y}}_{HH}$ denote the less redundant ("cleaned") LL, HL, LH and HH bands. The dashed lines indicate that the operator $\mathcal{T}^W_{H2L}$ is only optional.

In our previous work [30], we proposed a simple, yet effective, hand-tuned solution for the operator $\mathcal{T}^A_{L2H}$, which explicitly targets the discovery of local geometric flow in the low-pass band to untangle aliasing within the detail bands. This is certainly not the only way to design $\mathcal{T}^A_{L2H}$, and redundant information might be exploited in a more general way. The purpose of [30] is to demonstrate that an algorithm designed exclusively to exploit geometric flow, without statistical modeling or learning, is capable of untangling redundant information within the detail subbands.

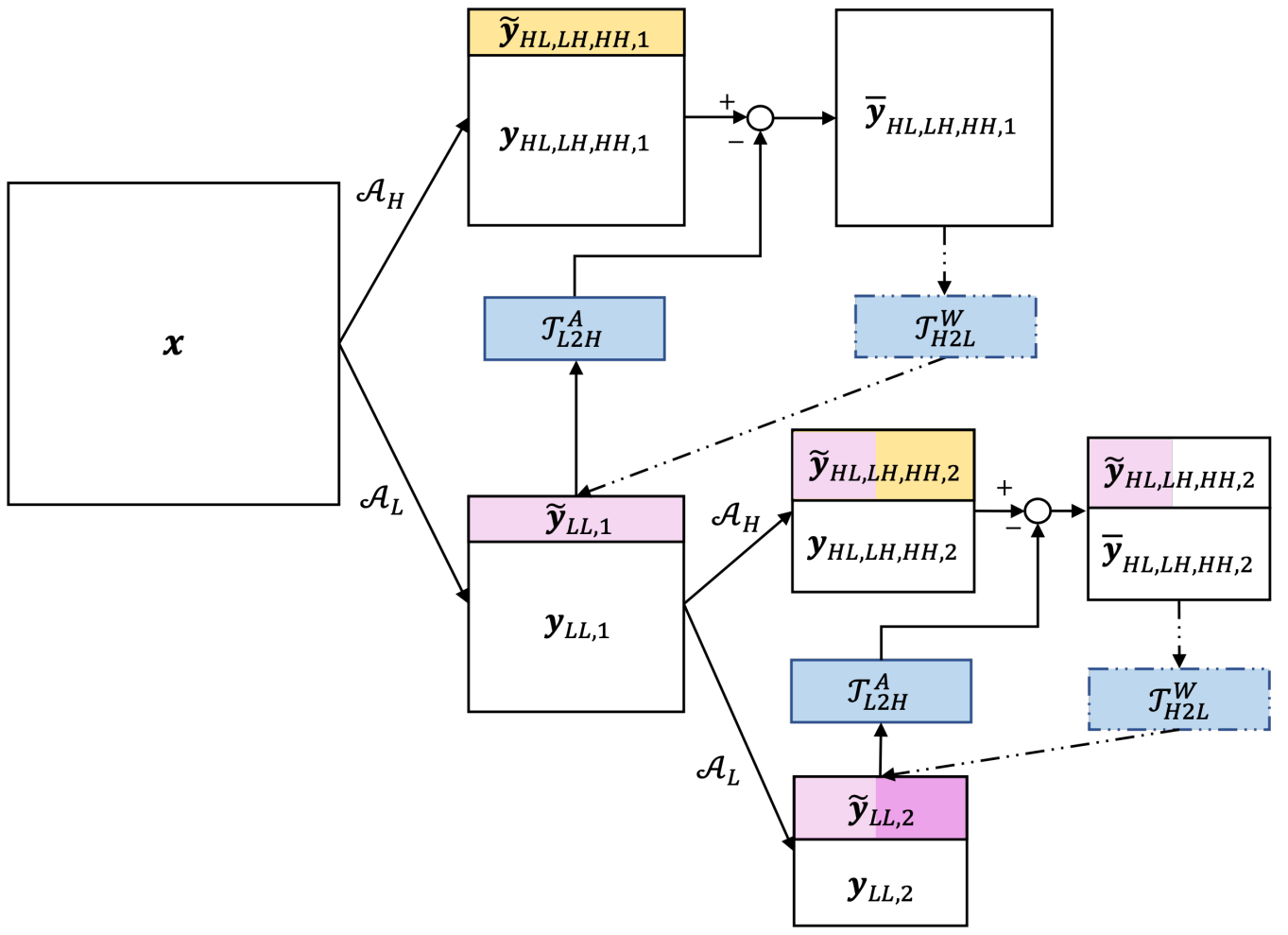

Fig. 3. Extending the low-to-high approach to coarser levels, where $\mathbf{y}_{LL,d}$, $\mathbf{y}_{HL,d}$, $\mathbf{y}_{LH,d}$ and $\mathbf{y}_{HH,d}$ represent the low- and high-pass bands at the $d^{th}$ level of decomposition. The symbols $\tilde{\mathbf{y}}_{LL,d}$, $\tilde{\mathbf{y}}_{HL,d}$, $\tilde{\mathbf{y}}_{LH,d}$ and $\tilde{\mathbf{y}}_{HH,d}$ denote the aliasing information within the low- and high-pass bands at level $d$. The symbols $\bar{\mathbf{y}}_{HL,d}$, $\bar{\mathbf{y}}_{LH,d}$ and $\bar{\mathbf{y}}_{HH,d}$ stand for the less redundant detail bands after applying the operator $\mathcal{T}^A_{L2H}$. The dashed lines indicate that the operator $\mathcal{T}^W_{H2L}$ is only optional.

Unfortunately, this approach does not extend well to coarser levels in the wavelet decomposition. The reason for

1. We have done this to help clarify the connection with super-resolution, which is always understood as starting from a low resolution image.

this can be understood with the aid of Fig. 3. We see the LL band at the first level of decomposition ($\mathbf{y}_{LL,1}$) cannot be regarded as samples of a continuous Nyquist band-limited image, as it contains the aliasing component $\widetilde{\mathbf{y}}_{LL,1}$ due to down-sampling. This aliasing component then accumulates through the DWT hierarchy, and forms part of the LL band at the next level of decomposition ($\mathbf{y}_{LL,2}$). Given the increasing amount of aliasing presented in $\mathbf{y}_{LL,2}$, it becomes harder to discover local properties such as geometric flow, reducing the effectiveness with which redundancy can be suppressed within the detail bands $\mathbf{y}_{HL,2}$, $\mathbf{y}_{LH,2}$ and $\mathbf{y}_{HH,2}$.

In the light of this fundamental difficulty, we choose not to pursue the development of more sophisticated low-to-high approaches. Instead, we propose to adopt a high-to-low approach, which uses the high-pass subbands $\mathbf{y}_{HL,d}$, $\mathbf{y}_{LH,d}$ and $\mathbf{y}_{HH,d}$ to remove redundant aliasing $\widetilde{\mathbf{y}}_{LL,d}$ from $\mathbf{y}_{LL,d}$ at each level $d$, before proceeding to the next level in the decomposition. We do this using an operator $\mathcal{T}^{A}_{H2L}$ as seen in Fig. 4. Similar to $\mathcal{T}^{A}_{L2H}$, we expect the operator $\mathcal{T}^{A}_{H2L}$ to also be capable of adaptively exploiting local geometric features from the detail bands to predict aliasing within the LL band. Conceptually, if the operator $\mathcal{T}^{A}_{H2L}$ successfully targets aliasing untangling within the LL band, then further reducing redundancy with the detail bands could be achieved simply using a linear operator $\mathcal{T}^{W}_{L2H}$ as explained in Section 2.

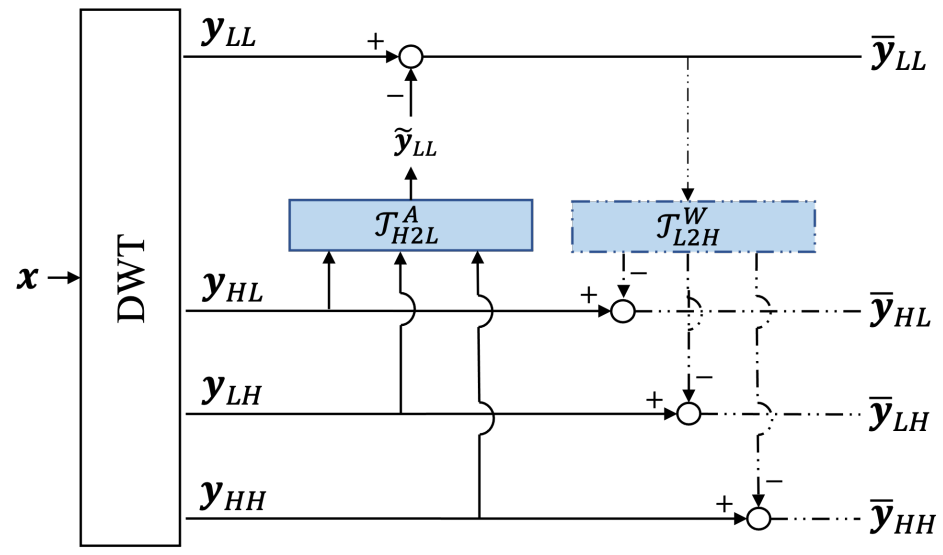

Fig. 4. The architecture of the high-to-low approach. The symbols $\mathbf{y}_{LL}$, $\mathbf{y}_{HL}$, $\mathbf{y}_{LH}$ and $\mathbf{y}_{HH}$ represent the LL, HL, LH and HH bands of the wavelet transform. The symbols $\bar{\mathbf{y}}_{LL}$, $\bar{\mathbf{y}}_{HL}$, $\bar{\mathbf{y}}_{LH}$ and $\bar{\mathbf{y}}_{HH}$ denote the less redundant ("cleaned") LL, HL, LH and HH bands. The dashed lines indicate that the operator $\mathcal{T}^{W}_{L2H}$ is only optional.

Contrary to the low-to-high approach, the high-to-low approach is expected to be more successful at untangling redundancy within the LL band; the accumulation of aliasing is then effectively avoided through the DWT hierarchy, which makes the method applicable to multiple levels of decomposition as seen in Fig. 5. Moreover, by effectively cleaning aliasing within the LL band at each level, reconstructed images at different scales indeed turn out to have significantly higher visual quality than the LL bands from the wavelet transform.

To develop the operator $\mathcal{T}^{A}_{H2L}$, preliminary experiments have been conducted for the high-to-low method using the hand-tuned solution presented in [30], which was not very successful. This is because it is more difficult to discover local geometric flow from the detail bands than from the low-pass band, at least without the aid of strong prior statistical models. For this reason, it seems appropriate to adopt machine learning as a tool for the methods presented in this section. More details concerning the proposed neural network structures themselves are presented in Section 4, but here we focus on architectural aspects.

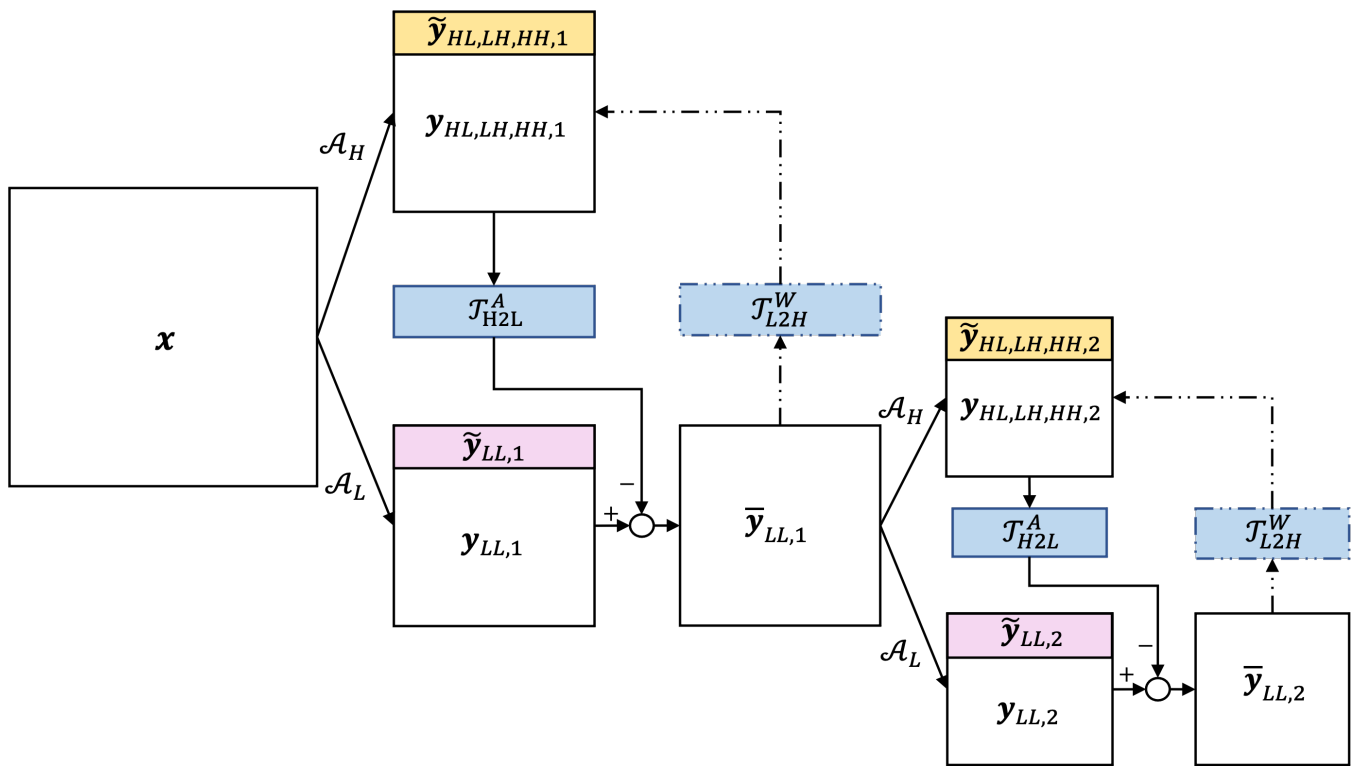

Fig. 5. Extending the high-to-low approach to lower levels, where $\widetilde{\mathbf{y}}_{LL,d}$, $\widetilde{\mathbf{y}}_{HL,d}$, $\widetilde{\mathbf{y}}_{LH,d}$ and $\widetilde{\mathbf{y}}_{HH,d}$ denote the aliasing information within the wavelet subbands at level $d$. The symbol $\bar{\mathbf{y}}_{LL,d}$ stands for the less redundant ("cleaned") LL band at level $d$ after applying the operator $\mathcal{T}^{A}_{H2L}$. The dashed lines indicate that the operator $\mathcal{T}^{W}_{L2H}$ is only optional.

Building on the high-to-low approach, we introduce a third "hybrid" architecture to further improve coding efficiency. Rather than employing a linear low-to-high operator $\mathcal{T}^{W}_{L2H}$ as described in the high-to-low approach, the hybrid architecture adopts an adaptive low-to-high operator $\mathcal{T}^{A}_{L2H}$ after implementing $\mathcal{T}^{A}_{H2L}$ as seen in Fig. 6. Although conceptually $\mathcal{T}^{W}_{L2H}$ is sufficient to suppress redundancy within the detail bands, it is strictly true only if the first operator $\mathcal{T}^{A}_{H2L}$ pre-cleans all aliasing from the LL band. By introducing an adaptive low-to-high operator, the hybrid approach can maintain the benefits of coding efficiency even if $\mathcal{T}^{A}_{H2L}$ fails to clean aliasing from the low-pass band in the first place.

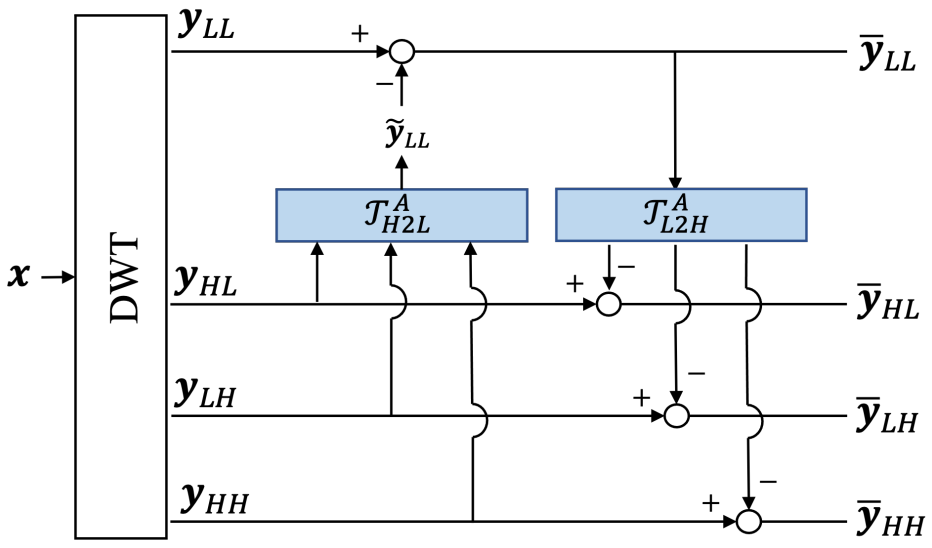

Fig. 6. The architecture of the hybrid method. The symbols $\mathbf{y}_{LL}$, $\mathbf{y}_{HL}$, $\mathbf{y}_{LH}$ and $\mathbf{y}_{HH}$ represent the LL, HL, LH and HH bands of the wavelet transform. The symbols $\bar{\mathbf{y}}_{LL}$, $\bar{\mathbf{y}}_{HL}$, $\bar{\mathbf{y}}_{LH}$ and $\bar{\mathbf{y}}_{HH}$ denote the less redundant ("cleaned") LL, HL, LH and HH bands.

In terms of the encoding system, it can be implemented in either open-loop or closed-loop fashion. The difference between the two approaches rests in how quantization errors are treated and propagated in the synthesis step. The details of each encoding approach are given below.

### 3.1 Closed-loop encoding system

The closed-loop encoding approach is conceptually appealing in the context of non-linear operators; it avoids the propagation of quantization errors, which otherwise are expanded in an uncontrollable way through non-linearities in the networks. To achieve this, the closed-loop encoding system essentially embeds the decoder inside the encoder, so that the transform is designed at the decoder with quantized data.

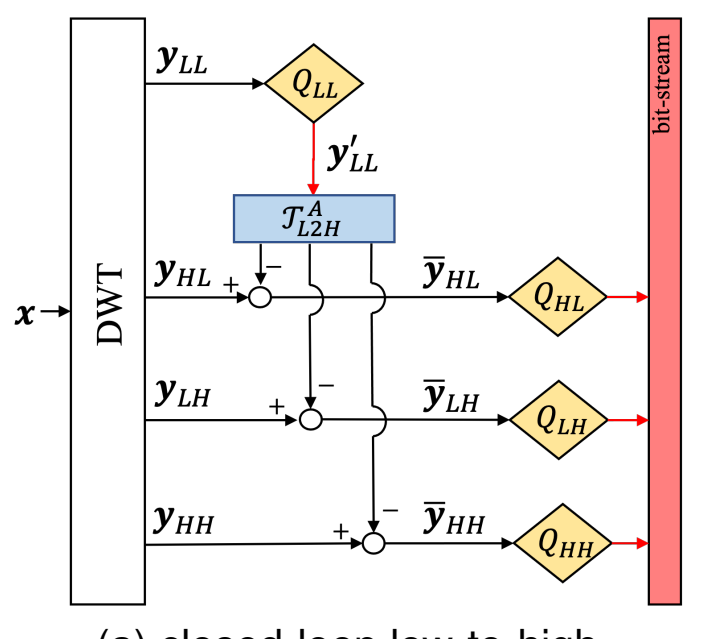

(a) closed-loop low-to-high

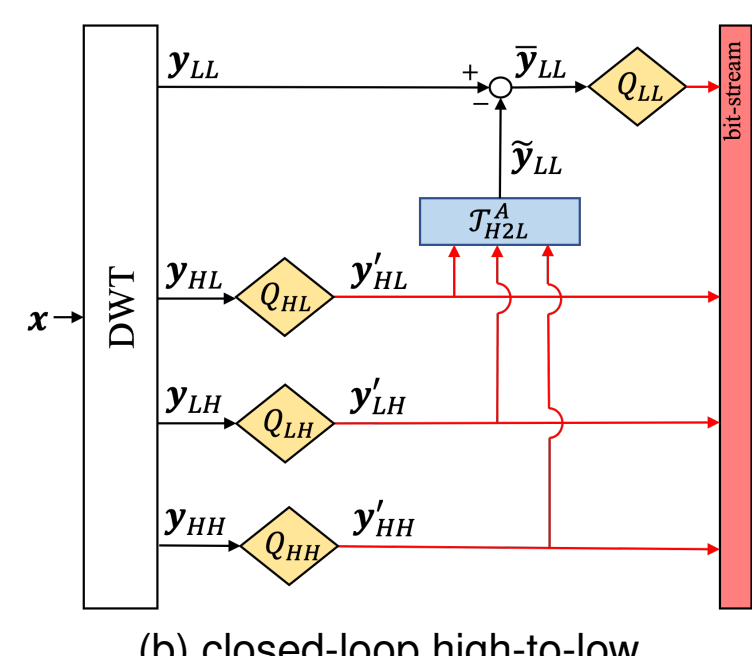

(b) closed-loop high-to-low

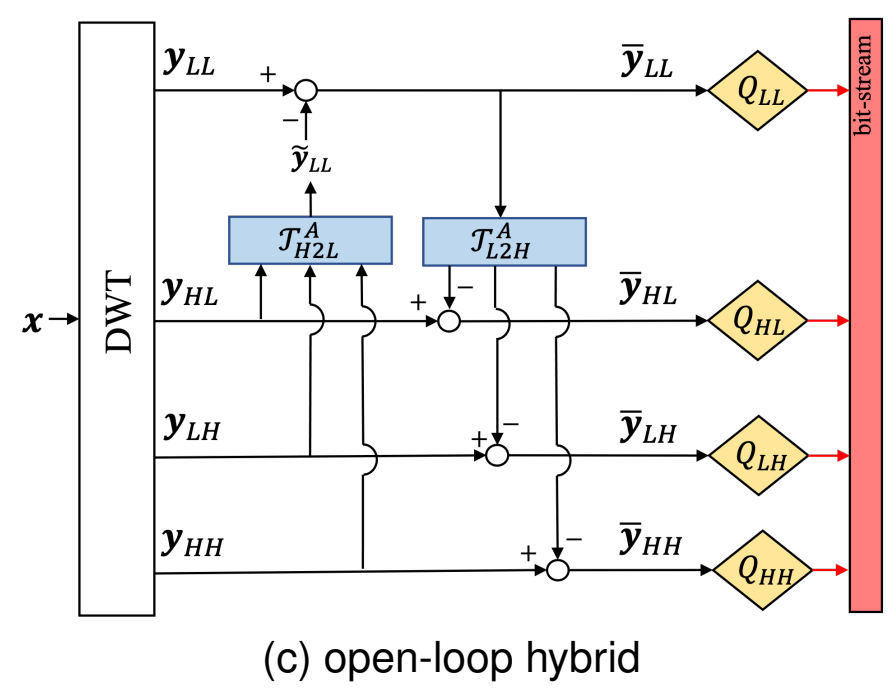

(c) open-loop hybrid

Fig. 7. The proposed closed-loop (a)(b) and open-loop (c) encoding frameworks. The symbols $Q_{LL}$, $Q_{HL}$, $Q_{LH}$ and $Q_{HH}$ represent the quantizer for the LL, HL, LH and HH bands respectively. The symbols $\mathbf{y}'_{LL}$, $\mathbf{y}'_{HL}$, $\mathbf{y}'_{LH}$ and $\mathbf{y}'_{HH}$ denote the quantized wavelet subbands.

In our scenario, the low-to-high and the high-to-low architectures can be developed respectively in the closed-loop encoding framework as seen in Fig. 7 (a) and (b). In both cases, adding additional Wiener filters $\mathcal{T}^{W}_{L2H}$ and $\mathcal{T}^{W}_{H2L}$ is infeasible, as it creates cyclic dependencies between the adaptive operators $\mathcal{T}^{A}_{L2H}$ and $\mathcal{T}^{A}_{H2L}$; this prevents us from finding a deterministic process for determining the quantized subband samples. For this same reason, the closed-loop encoding system is incompatible with the hybrid architecture. Considering these fundamental difficulties, we choose to focus on developing the open-loop encoding system as presented in the following section.

### 3.2 Open-loop encoding system

In the so-called "open-loop" approach, the transform is designed at the encoder without any quantization, whereas the decoder receives quantized samples to invert the operation. In this scenario, the hybrid architecture is feasible, as illustrated in Fig. 7(c), which is of particular interest due to its ability to adaptively remove redundancy within both the $\mathcal{T}^{A}_{H2L}$ and $\mathcal{T}^{A}_{L2H}$ steps.

The main challenge for open-loop encoding is that quantization errors propagate through multiple adaptive operators that necessarily entail non-linear elements, in addition to the linear wavelet synthesis operators themselves. This introduces the potential for quantization errors to be amplified, in ways that are strongly data dependent and hence harder to bound. Ultimately, this will require careful modeling during the training of our neural network based operators. Nonetheless, it turns out that it is possible to develop open-loop hybrid architectures that achieve significant gains in coding efficiency across a wide range of bit-rates, in a completely scalable setting.

## 4 Neural Network Structures

In Section 2, we have mathematically elaborated the opportunity that exists to exploit the residual redundancy from the existing wavelet transform, i.e. a simple linear solution is sufficient to untangle the redundant (notably aliasing) information within regions with consistent geometric flow. The purpose of this section is to give insight on how this underlying hypothesis drives the structure of the neural networks that we select. Eventually, we find that the best solution does indeed involve banks of optimized linear filters controlled dynamically by an opacity network. This confirms our underlying hypothesis that the solution to our problem (redundancy exploitation) can be a linear filter if the local orientation is known a priori.

In a preliminary exploration phase, we explore the merits of different structures. This exploration phase does not involve end-to-end training for the full rate-distortion optimization problem. Instead, we measure the energy compaction potential of different structures and we explore robustness to quantization error propagation by considering just one level of decomposition in isolation. Later, after identifying the most suitable structures, we develop a comprehensive end-to-end training strategy that is capable of modeling the complex interactions between quantization and adaptive processing steps across the decomposition hierarchy; this is the subject of Section 5.

### 4.1 Benefits of proposal-opacity structures

Our exploration phase starts with a focus only on the adaptive high-to-low operator $\mathcal{T}^{A}_{H2L}$. This is because $\mathcal{T}^{A}_{H2L}$ is the most critical element to avoid propagation of aliasing through the DWT hierarchy, and opens the opportunity for the transform architecture to be extended to multiple wavelet decomposition levels. This is surely not the only way to approach the problem, but the initial explorations involving only $\mathcal{T}^{A}_{H2L}$ turn out to be very insightful.

We begin by considering a fairly straightforward high-to-low network structure in [31]. This structure is composed from three subnetworks involving conventional convolution and *Leaky ReLU* operators, as seen in Fig. 8. Variations on this structure were also explored, involving concatenation of the HL, LH and HH source channels ahead of the first convolution layer.

To evaluate the potential of these high-to-low network structures without building a complete end-to-end optimization system, our primary training objective is aliasing suppression within the LL bands. This objective is chosen for two reasons: 1) removal of aliasing is necessary to ensure that the approach can be effectively applied also at coarser levels in the DWT hierarchy, as elaborated in Fig. 5; and 2) aliasing suppression also helps to reduce redundancy from the subbands that are derived from the "cleaned" LL band. This naturally leads to higher energy compaction, which can be employed as an evaluation criterion when assessing the performance of these high-to-low networks.

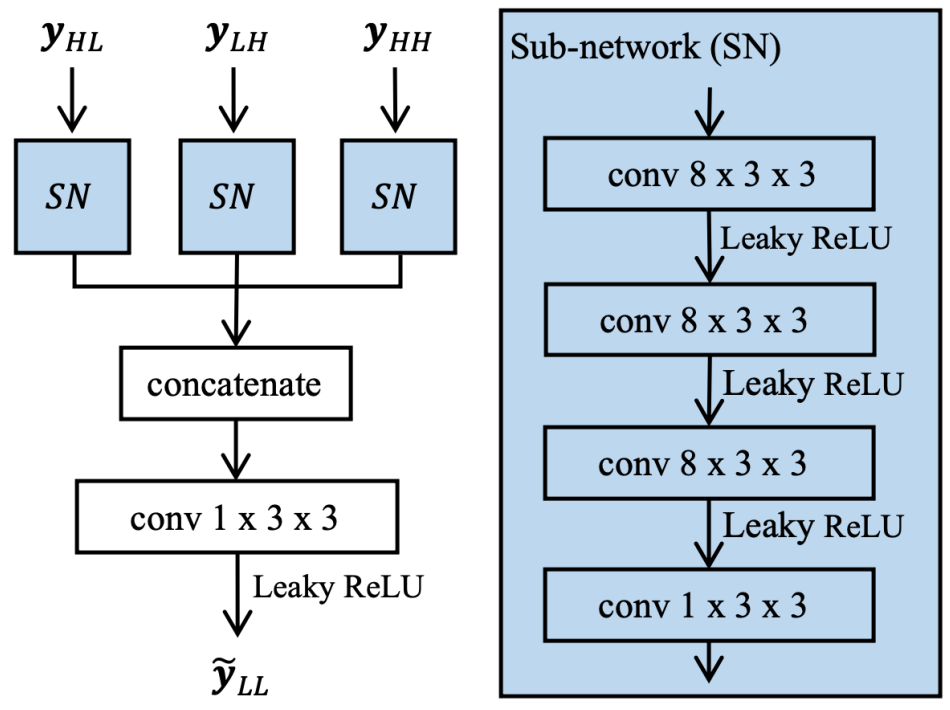

Fig. 8. The initial high-to-low network structure proposed in [31], where N x K x K denotes N filters (or channels) with kernel support K x K.

To be more specific, the idea of developing our training objective is to construct a model $\tilde{\mathbf{y}}^t_{LL,d}$ for the aliasing in the LL band at each level of decomposition $d$ by subtracting $\mathbf{y}_{LL,d}$ from $\bar{\mathbf{y}}^t_{LL,d}$ as seen in Fig. 9. The accent $^-$ is used to indicate the subband free of aliasing, while the superscript $^t$ denotes the training target. The subband $\bar{\mathbf{y}}^t_{LL,d}$ is obtained by low-pass filtering $\bar{\mathbf{y}}^t_{LL,d-1}$ and then subjecting it to the low-pass wavelet analysis operator $\mathcal{A}_L$, while $\mathbf{y}_{LL,d}$ is derived from the "cleaned" LL band $\bar{\mathbf{y}}_{LL,d-1}$ (or the image $\mathbf{x}$) by applying the low-pass wavelet operator $\mathcal{A}_L$. The low-pass filter (LPF) employed here has a windowed sinc impulse response with bandwidth $0.7\pi$.

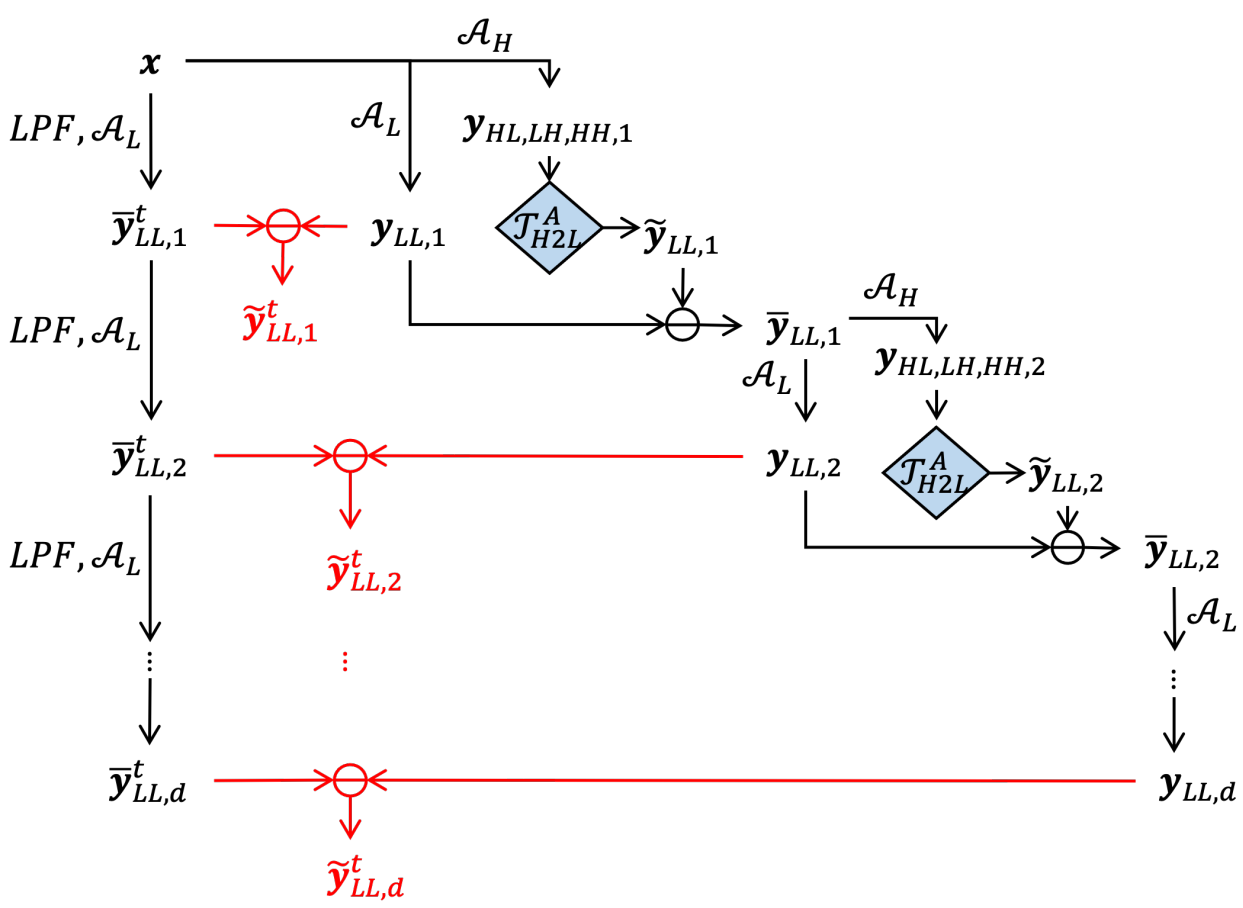

Fig. 9. The proposed structure to construct the aliasing model $\tilde{\mathbf{y}}^t_{LL,d}$ for the LL band at each decomposition level $d$.

The objective function can be either the $l_2$-norm $\left\|\tilde{\mathbf{y}}_{LL,d} - \tilde{\mathbf{y}}^t_{LL,d}\right\|_2^2$ or the $l_1$-norm $\left\|\tilde{\mathbf{y}}_{LL,d} - \tilde{\mathbf{y}}^t_{LL,d}\right\|_1$, where $\tilde{\mathbf{y}}_{LL,d}$ is the aliasing predicted by the high-to-low operator (network) $\mathcal{T}^A_{H2L}$. The difference between these two objective metrics will not be explicitly addressed, as we have empirically verified that their impacts on the performance of different high-to-low networks is neglectable.

For the experimental results, the Adam algorithm [32] is employed for training, with 75 image batches comprising 16 patches of size 256 x 256 from the DIV2K image dataset. Other images in DIV2K dataset that are not included during training are used for testing. To evaluate the performance of different high-to-low network structures, we consider two objective measurements: 1) energy compaction, that is the ratio of the energy of the original detail bands obtained through LeGall 5/3 wavelet transform to the detail bands $\mathbf{y}_{HL,d}$, $\mathbf{y}_{LH,d}$ and $\mathbf{y}_{HH,d}$ decomposed from the "cleaned" LL band $\bar{\mathbf{y}}_{LL,d-1}$; and 2) visual enhancement of the "cleaned" LL band ($\bar{\mathbf{y}}_{LL,d}$) at different resolutions $d$.

TABLE 1

(a) Energy compaction of the initial network structure proposed in [31] as seen in Fig. 8.

| | LL | HL | LH | HH |
|---|---|---|---|---|
| level 1 | 99.7% | – | – | – |
| level 2 | 99.9% | 91.2% | 88.9% | 76.7% |
| level 3 | 99.9% | 96.4% | 93.5% | 83.4% |
| level 4 | 100.5% | 97.5% | 94.8% | 85.1% |
| level 5 | 100.8% | 102.4% | 96.6% | 93.3% |

(b) Energy compaction of the proposal-opacity network structure with linear proposals seen in Fig. 11.

| | LL | HL | LH | HH |
|---|---|---|---|---|
| level 1 | 99.5% | – | – | – |
| level 2 | 99.6% | 88.4% | 85.8% | 68.2% |
| level 3 | 99.0% | 94.8% | 91.2% | 74.7% |
| level 4 | 98.4% | 90.4% | 86.4% | 69.8% |
| level 5 | 99.1% | 86.9% | 87.9% | 65.1% |

(c) Energy compaction of the proposal-opacity structure with non-linear proposals seen in Fig. 11.

| | LL | HL | LH | HH |
|---|---|---|---|---|
| level 1 | 99.4% | – | – | – |
| level 2 | 99.3% | 85.3% | 83.7% | 64.4% |
| level 3 | 99.0% | 90.4% | 87.1% | 67.2% |
| level 4 | 98.7% | 90.6% | 85.2% | 67.8% |
| level 5 | 99.5% | 89.0% | 88.0% | 65.9% |

Table 1(a) provides numerical results to illustrate the averaged energy compaction of the initial high-to-low network structure shown in Fig. 8 across all images in the testing set. In the experiment, we employ 5 levels of the LeGall 5/3 bi-orthogonal DWT, applying the proposed neural network prediction strategy for all levels of decomposition. As we see, the energy compaction of the detail subbands at all the levels affected by the operator $\mathcal{T}^A_{H2L}$ can be reduced considerably; those levels not affected by $\mathcal{T}^A_{H2L}$ are identified by a "–" in Table 1. The visual enhancement of the "cleaned" LL band obtained from this simple structure can be found in Fig. 10(b). Variations on this network structure are not explicitly shown here, as they empirically show similar potential to the initial one seen in Fig. 8.

Although this initial network structure appears to work, the underlying theory presented in Section 2 suggests that it should be possible to develop a linear solution to untangle redundant (aliasing) information within regions where local geometric flow is consistent. This reasoning suggests that we would do well to decompose the high-to-low network in two aspects: a bank of learned linear filters, each capable

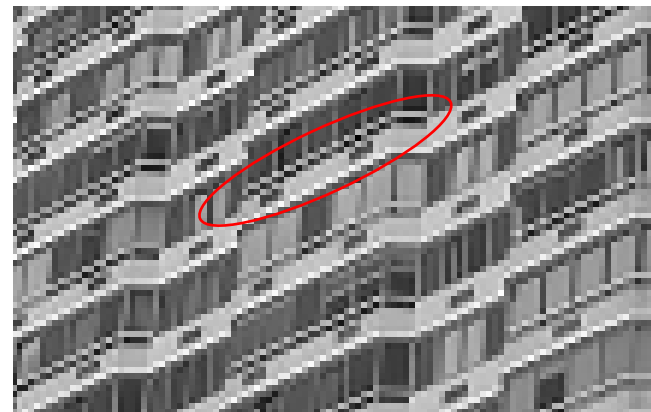

(a) original LL band, with aliasing ("staircases") along edges.

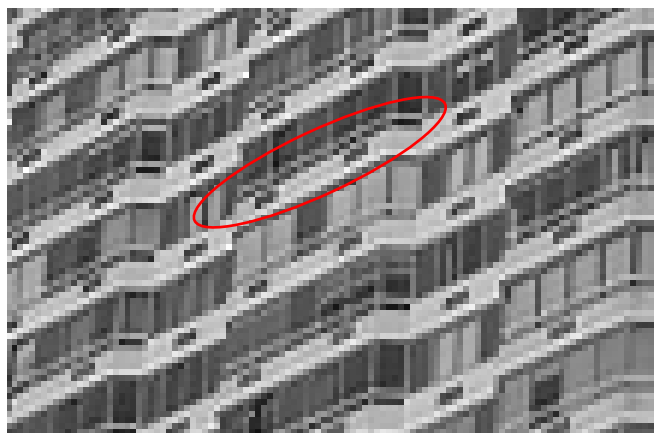

(b) the initial structure proposed in [31], with much less aliasing

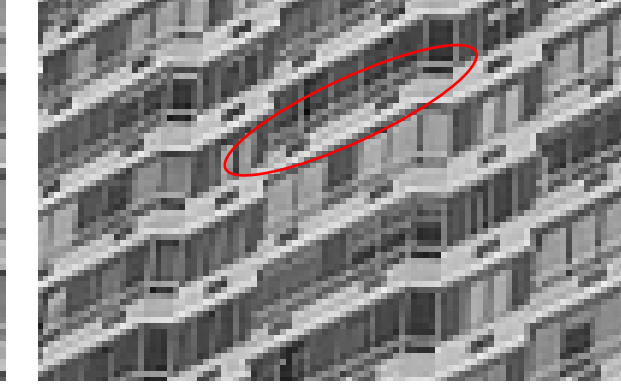

(c) proposal-opacity structure with linear proposals and sigmoid

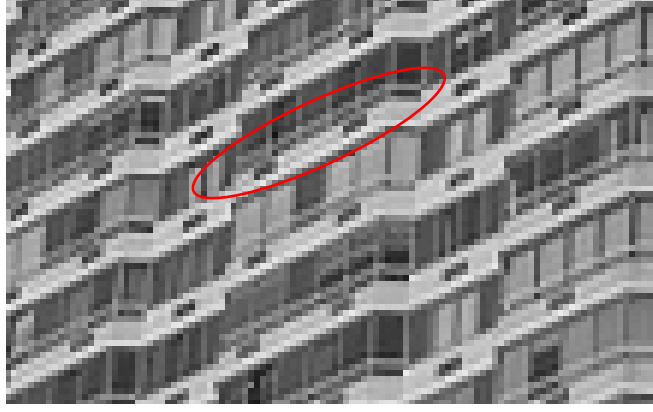

(d) proposal-opacity structure with the non-linear proposal network

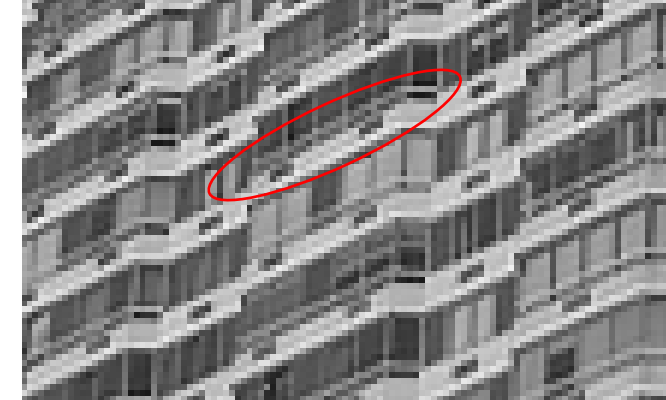

(e) proposal-opacity structure with linear proposals, log-like activation

Fig. 10. Visual quality of the "cleaned" LL bands at the third finest resolution from different network structures. We are specifically looking for aliasing suppression, i.e. less staircase-like artifacts around edges.

of responding to different geometric features; and a separate feature detector network, which is necessarily non-linear.

Specifically, we explore proposal-opacity structures, as shown in Fig. 11, where the non-linear opacity network ($N = 8$) is understood as analyzing local scene geometry to produce opacities (or likelihoods) in the range 0 to 1 that are used to blend linearly generated proposals for the aliasing prediction term. The structure of the opacity network is inspired by [33], employing residual blocks that have been demonstrated to be useful in feature detection, while the proposals are chosen to have the same region of support as the opacity network. Since the proposals are completely linear, if our training objective is the $l_2$-norm $\left\|\widetilde{\mathbf{y}}_{LL,d} - \widetilde{\mathbf{y}}^{t}_{LL,d}\right\|_2^2$, the proposal system amounts to a linear least mean-squared error (LLMSE) best estimator conditioned on the opacities, so it is effectively a bank of Wiener filters.

By comparing the energy compaction in Table 1(a) and (b), it can be seen that the proposal-opacity network structure indeed achieves considerably higher energy compaction for all the relevant detail bands across all the levels. Moreover, this proposal-opacity structure does produce more visually meaningful LL bands at different resolutions with less "staircases" around edges, compared with that of the LeGall 5/3 wavelet transform and the initial structure in

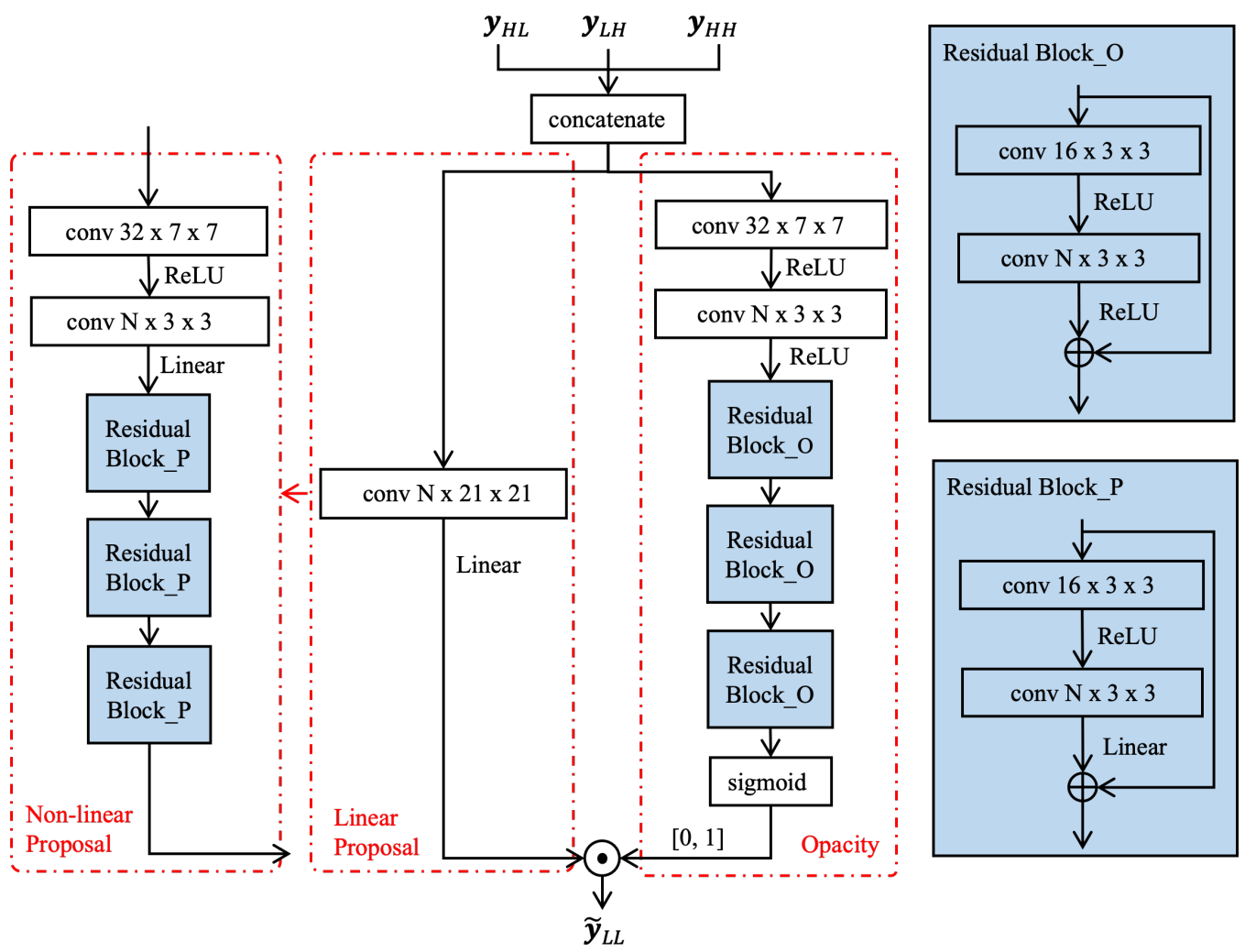

Fig. 11. The proposed proposal-opacity structure for the high-to-low network, with the linear or non-linear proposals and sigmoid as activation; N x K x K denotes N filters (or channels) with kernel support K x K.

Fig. 8; see examples in Fig. 10(a)(b)(c).

### 4.2 Sufficiency of linear proposal structures

It is worth considering whether the proposal-opacity structure can be improved by introducing nonlinearities into the proposal network as well. Specifically, we choose the proposal network ($N = 8$) to be substantially similar to the opacity network, as depicted in the left of Fig. 11, whereas *ReLU* and linear activation functions alternate to ensure zero-mean outputs. By comparing the prediction effectiveness in Table 1(b) and (c) and the visual quality of the LL bands in Fig. 10(c) and (d), the linear proposal structure seems to have comparable performance to the non-linear one for the high-to-low operator $\mathcal{T}^{A}_{H2L}$.

To gain further insight into the benefits of linear versus non-linear proposal structures, we construct a complete hybrid architecture by extending the proposal-opacity concept to the low-to-high network $\mathcal{T}^{A}_{L2H}$, with linear or non-linear proposals ($N = 8$) as seen in Fig. 13. As a result, the open-loop coding efficiency can now be explored, instead of using energy compaction as a proxy, to understand the potential of different network structures.

In this open-loop setting, both $\mathcal{T}^{A}_{H2L}$ and $\mathcal{T}^{A}_{L2H}$ are trained without incorporating any quantization errors during training. $\mathcal{T}^{A}_{H2L}$ explicitly targets the aliasing model $\widetilde{\mathbf{y}}^{t}_{LL,d}$ during training as depicted in Fig. 9. $\mathcal{T}^{A}_{L2H}$ is trained to minimize the prediction residuals of the detail bands at each level $d$; that is either the $l_1$-norm $\left\|\mathbf{y}_{HL,d} - \widetilde{\mathbf{y}}_{HL,d}\right\|_1$ or $l_2$-norm $\left\|\mathbf{y}_{HL,d} - \widetilde{\mathbf{y}}_{HL,d}\right\|_2^2$, as depicted in Fig. 12. Although either objective function can be used to train $\mathcal{T}^{A}_{L2H}$, we have empirically verified that the $l_1$-norm results in higher open-loop coding efficiency. For simplicity, $\mathcal{T}^{A}_{H2L}$ is trained first, after which $\mathcal{T}^{A}_{L2H}$ is trained while keeping $\mathcal{T}^{A}_{H2L}$ fixed.

From Fig. 14 we can see that by applying $\mathcal{T}^{A}_{H2L}$ and $\mathcal{T}^{A}_{L2H}$ to only the finest resolution in the open-loop setting, the linear proposal structure is actually better than the non-linear one in terms of rate-distortion performance. This empirically confirms that a classic set of Wiener filters attenuated by corresponding opacities (or likelihoods) is competitive with and even superior to a fully non-linear solution, which reinforces the theoretical arguments presented in Section 2.

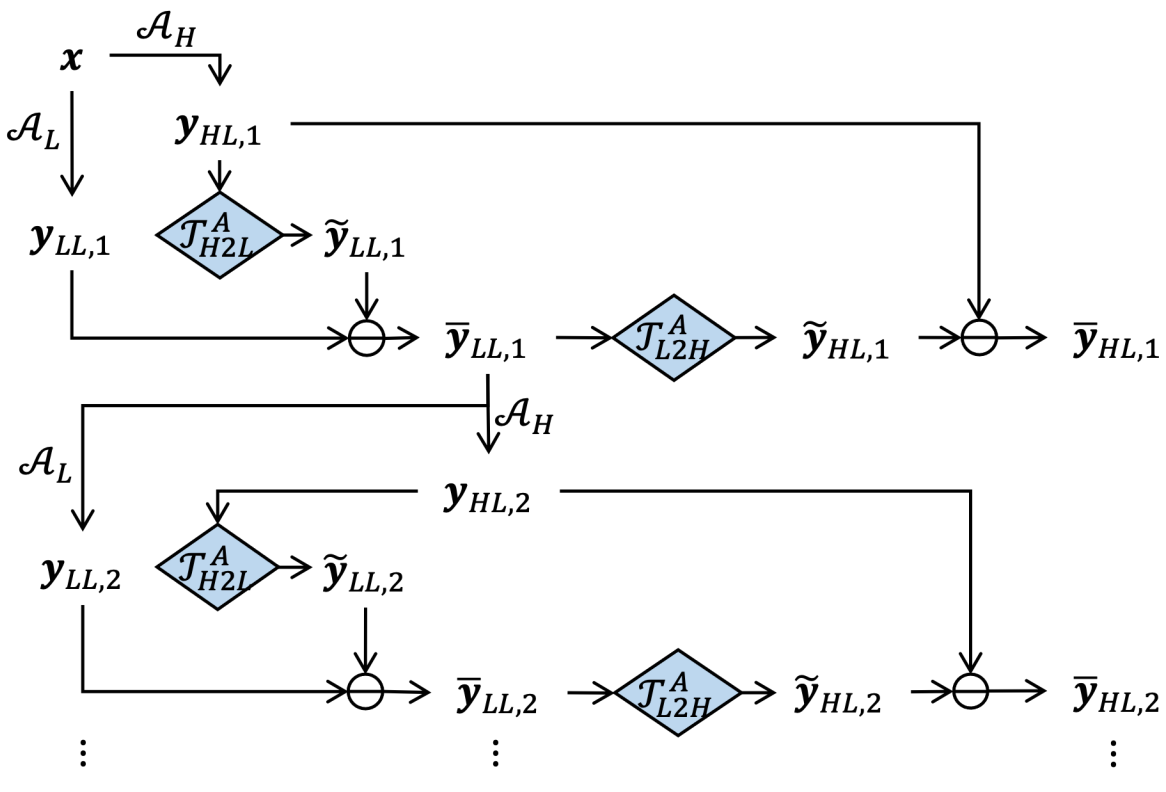

Fig. 12. The idea to generate the training objective for the low-to-high network. We use the HL band as an example here; the same methodology can be adopted for the LH and HH band.

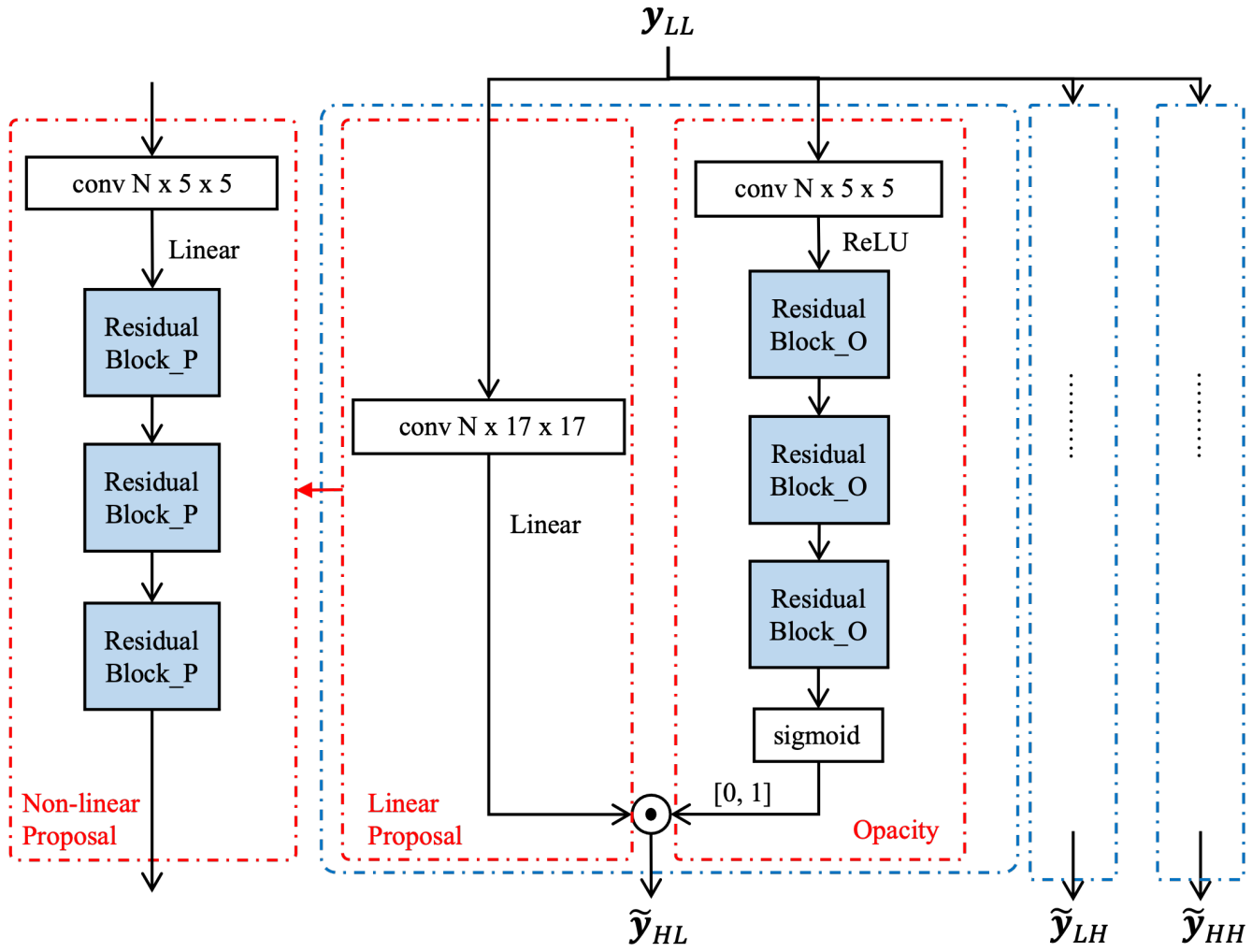

Fig. 13. The proposed proposal-opacity structure of the low-to-high network, with linear or nonlinear proposals and sigmoid as activation; N x K x K denotes N filters (or channels) with kernel support K x K.

### 4.3 Appropriate activation functions

In this section, we consider the opacity network more carefully. Following the underlying theory elaborated in Section 2, we expect the opacity network to model geometric features in the scene, which should be invariant to absolute image intensity and contrast. Unfortunately, the conventional sigmoid activation shown in Fig. 11 and Fig. 13 does not have this property. We expect to do better, therefore, by replacing the sigmoid with a log-like activation function.

In particular, we evolve the sigmoid function into a block as seen in Fig. 15. The log-like function that we adopt in this block is

$$y = \begin{cases} \log\left(x + \text{offset}\right), & x > -\text{offset}/2 \\ \log\left(\text{offset}/2\right), & \text{otherwise} \end{cases}$$

where offset $= 0.01$ is chosen to define the derivative of the function at the origin. This log-like activation function is followed by a linear convolution layer, which is expected to choose the dominant geometric feature. In the end, *tanh* and *ReLU* are concatenated to cap the opacities within the range $[0, 1]$. Interestingly, we see the structure with the log-like activation function does perform better than that with the sigmoid function in the open-loop encoding system, even with fewer channels ($N = 4$). Meanwhile, the visual quality of the "cleaned" LL band is still maintained; see Fig. 10 and Fig. 14 for more details.

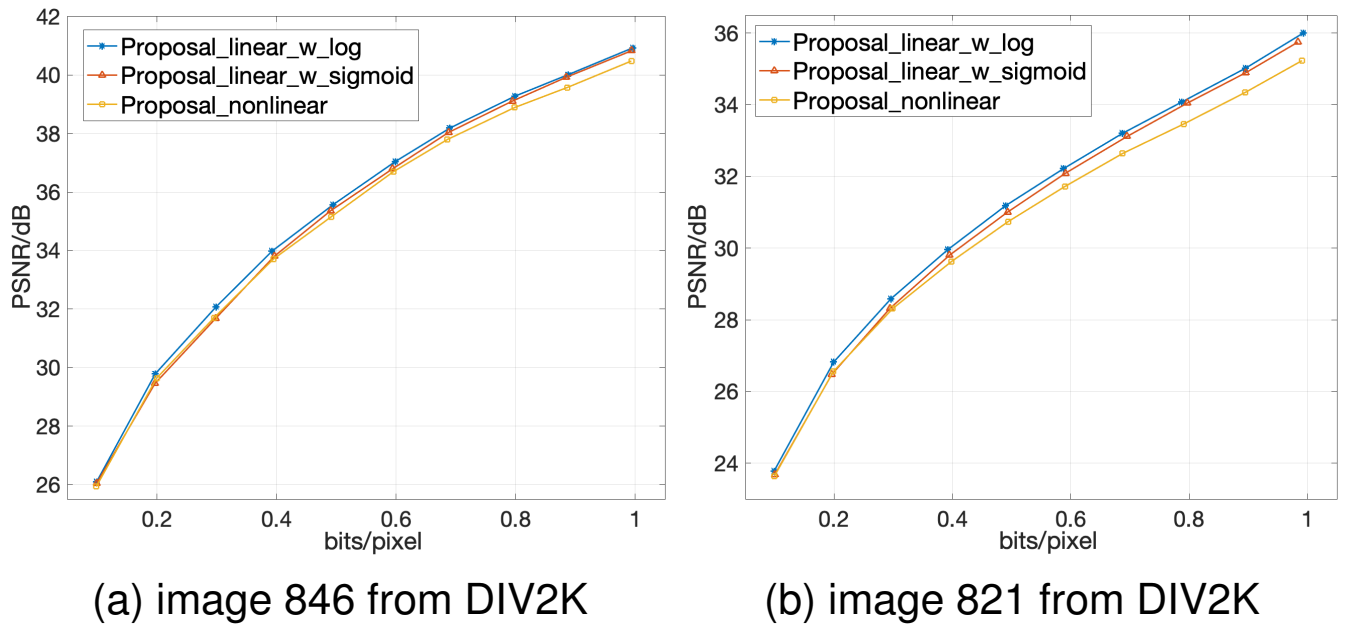

Fig. 14. The rate-distortion performance in the primitive open-loop setting for different proposal-opacity network structures: linear or non-linear proposals with sigmoid as the last activation function, as seen in Fig. 11 and Fig. 13; linear proposals with log-like activation as seen in Fig. 15.

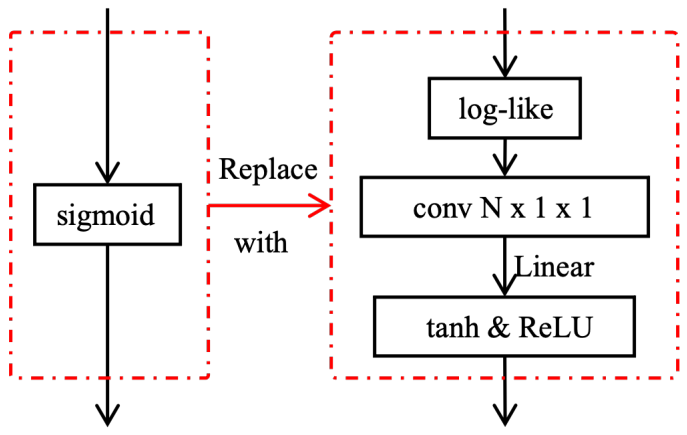

Fig. 15. The replacement of sigmoid with a log-like activation block.

## 5 Learning Strategy

In this section, we aim to jointly train the high-to-low and low-to-high networks for multiple levels of decomposition, along with the extra distortion gains introduced by these inference machines in addition to the base wavelet transform. The entire end-to-end optimization framework is depicted in Fig. 16. Interestingly, we eventually discover that a single pair of jointly trained high-to-low and low-to-high networks can be employed at all levels in the DWT decomposition hierarchy – that is, there is no need to learn and store separate network weights for each decomposition level.

Previously, aliasing suppression was our sole training objective in the initial exploration, since propagation of aliasing from high to low levels in the hierarchy would destroy the properties required for successful deployment of the approach at lower levels. As explained earlier, aliasing removal should be a reasonable proxy training objective when a single level of the hierarchy is considered in isolation. Now that we are embarking on an en-to-end learning strategy for $\mathcal{T}^A_{H2L}$ and $\mathcal{T}^A_{L2H}$, taking all levels of the hierarchy into account together, it is possible to replace our training objective with one that focuses exclusively on rate-distortion performance. As we shall see, however, the aliasing suppression objective is quite compatible with end-to-end rate-distortion optimization. To expose this fact, we retain an aliasing suppression term as one part of the training objective, which can be selectively included to explore the role it plays in our final solution.

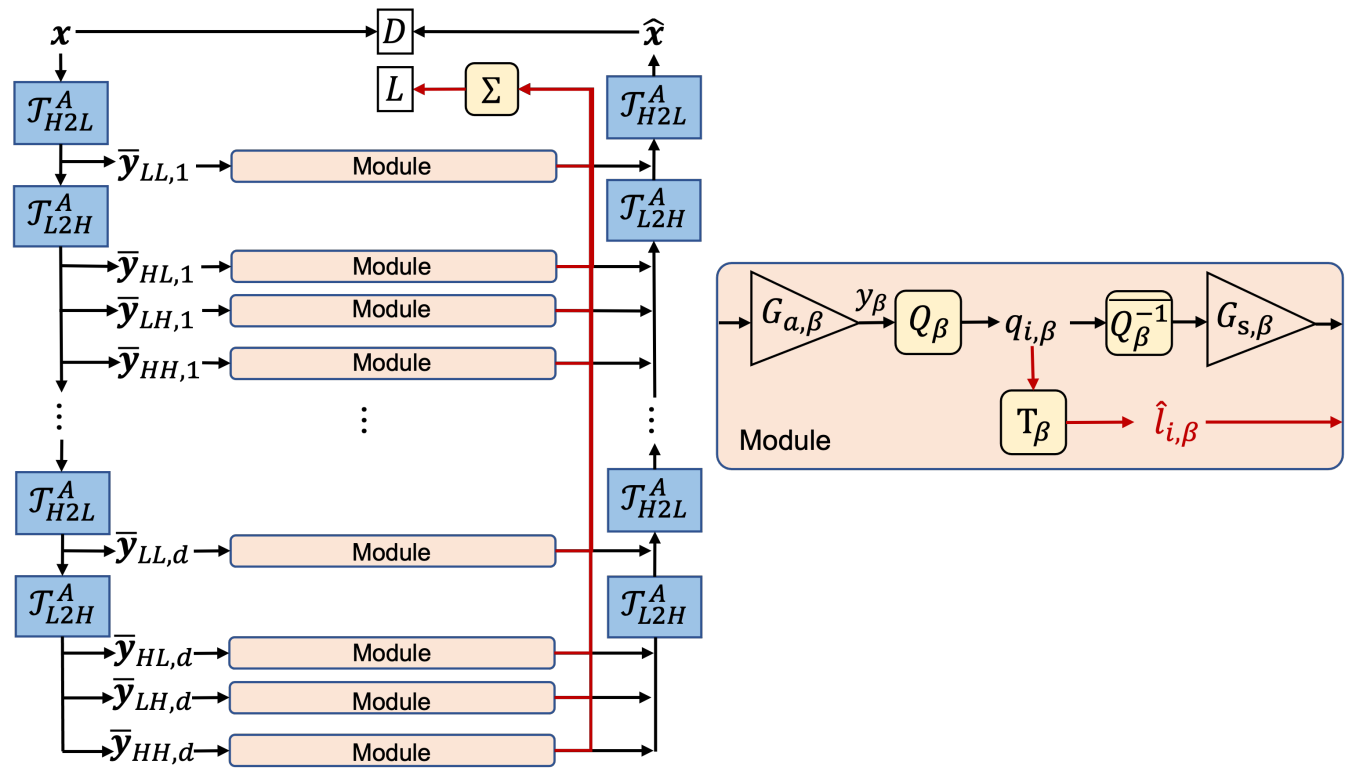

Fig. 16. The proposed end-to-end optimization framework. $G_{a,\beta}$ and $G_{s,\beta}$ denote the extra analysis and synthesis gains introduced by the neural networks in addition to the base wavelet transform for subband $B_\beta$. By evolving $G_{a,\beta}$ and $G_{s,\beta}$ during training, we effectively optimize the quantization step size of the quantizer $Q_\beta$ and the dequantizer $\overline{Q_\beta^{-1}}$ for subband $B_\beta$. Moreover, $q_{i,\beta}$ denotes the quantization indices $q_i$ within subband $B_\beta$, while $T_\beta$ is the look-up table that we use to map each $q_{i,\beta}$ to its respective coded length $\hat{l}_{i,\beta}$ for subband $B_\beta$.

To be more specific, our objective is to minimize:

$$J(\phi) = \underbrace{\|\mathbf{x} - \hat{\mathbf{x}}(\phi)\|^2}_{D} + \lambda_1 \underbrace{\sum_{\beta} \sum_{i \in B_\beta} l_{i,\beta}}_{L} + \lambda_2 \underbrace{\sum_{d} \left\| \widetilde{\mathbf{y}}_{LL,d}(\phi) - \widetilde{\mathbf{y}}^t_{LL,d} \right\|_2^2}_{\text{aliasing constraint term}} \tag{2}$$

where

$$l_{i,\beta} = \log_2 \frac{1}{P_{V_\beta}(q_{i,\beta};\phi)} = \log_2 \frac{1}{\mathrm{Prob}(V_\beta = q_{i,\beta};\phi)} \tag{3}$$

In (2), the total distortion term $D$ represents the sum of squared errors between the input image $\mathbf{x}$ and its reconstructed counterpart $\hat{\mathbf{x}}$; $\phi$ represents the vector of all network weights. The total coded length term $L$ is the sum of all coded lengths $l_{i,\beta}$, resulting from the coding of quantization indices $q_{i,\beta}$ for all subbands $B_\beta$. We write $V_\beta$ for the random variable from which the quantization indices $q_{i,\beta}$ are drawn; then, the coded length $l_{i,\beta}$ is modelled by (3). The LL band aliasing constraint term in (2) measures the sum of squared errors between $\widetilde{\mathbf{y}}_{LL,d}$ and $\widetilde{\mathbf{y}}^t_{LL,d}$ across all levels of decomposition $d$ as described in Section 4.1 and depicted in Fig. 9. The Lagrange multiplier $\lambda_1$ controls the trade-off between distortion $D$ and coded length $L$, while the other Lagrange multiplier $\lambda_2$ controls the level of emphasis on the visual quality of reconstructed images at different scales.

Eventually, we discover that constraining the aliasing term only at the finest resolution is sufficient for all intermediate resolutions to look good; this makes sense considering that we use the same set of network weights at all levels. In the training phase, we explore three settings of $\lambda_2$: 1) $\lambda_2 = 0$ to target rate-distortion performance alone; 2) $\lambda_2 = 1$ to encourage enhanced visual quality of LL bands within the rate-distortion optimization framework; and 3) $\lambda_2$ decreasing progressively from 1 to 0 through the training regime, so as to steer the training towards solutions that with visually appealing LL bands, while ultimately targeting rate-distortion performance alone.

To train this end-to-end optimized system for the objective in (2), most of the machine-learning optimization techniques, e.g. gradient decent, rely on differentiability for back-propagation. However, both the total distortion $D$ and the total coded length $L$ depend on the quantizer $Q_\beta$ and the dequantizer $\overline{Q_\beta^{-1}}$, whose derivatives are either zero or infinity everywhere. In the literature, many approaches have been proposed to resolve this issue. Ballè et al [16] propose to replace the quantizer with an additive uniform noise source; similar work can also be found in [34]. These additive noise models have the forward-backward discrepancy, because the forward-pass uses realizations of the noise model while the backward-pass employs the probability distribution (statistical model) of the noise. Moreover, these approaches also suffer from train-test discrepancy that the forward-pass with additive noise is different from the actual quantization that takes place in the real testing phase.

Another approach is to use a straight-through estimator (STE) [35] [36]; in this approach, the identity is employed as a continuous relaxation of quantization only for the backward pass of back-propagation. Since the forward-pass of the STE still adopts the non-differentiable quantization step function, it essentially avoids train-test discrepancy and has been proven to have a major benefit in performance over the additive noise approaches. However, it still suffers from the forward-backward mismatch, which has a fundamental impact on the convergence of the system.

In contrast, soft-to-hard annealing approaches are proposed to develop a continuous relaxation of quantization for both the forward- and the backward-pass during training [37] [38], so that there is no forward-backward mismatch. This continuous relaxation gets gradually annealed towards the actual non-differentiable quantization step function throughout the training to reduce the train-test discrepancy. However, the neural networks in these annealing methods do not have early access to the real quantized data that is actually used during testing.

To take advantage of these existing methods, we propose a backward annealing approach, which essentially interpolates the discontinuous function using a sliding Gaussian to form a continuous relaxation of the non-differentiable step functions in the backward pass, whereas the forward pass retains its original discontinuous quantization behavior, as seen in Fig. 17. In this way, our method avoids train-test mismatch, which however exists in additive noise approaches. By gradually reducing the standard deviation $\sigma$ of the sliding Gaussian during training, the fitness of the continuous relaxation to the true discontinuous operator can be easily annealed. This means that we can gradually eliminate the forward-backward discrepancy while still provide the networks an accurate visibility to real quantized data early on during training; this is in contrast to STE and soft-to-hard annealing approaches.

Specifically, assume we wish to develop a differentiable approximation function $\widetilde{Q}_\beta$ for the quantizer $Q_\beta$ of subband $B_\beta$, which is a uniform scalar quantizer with deadzone as

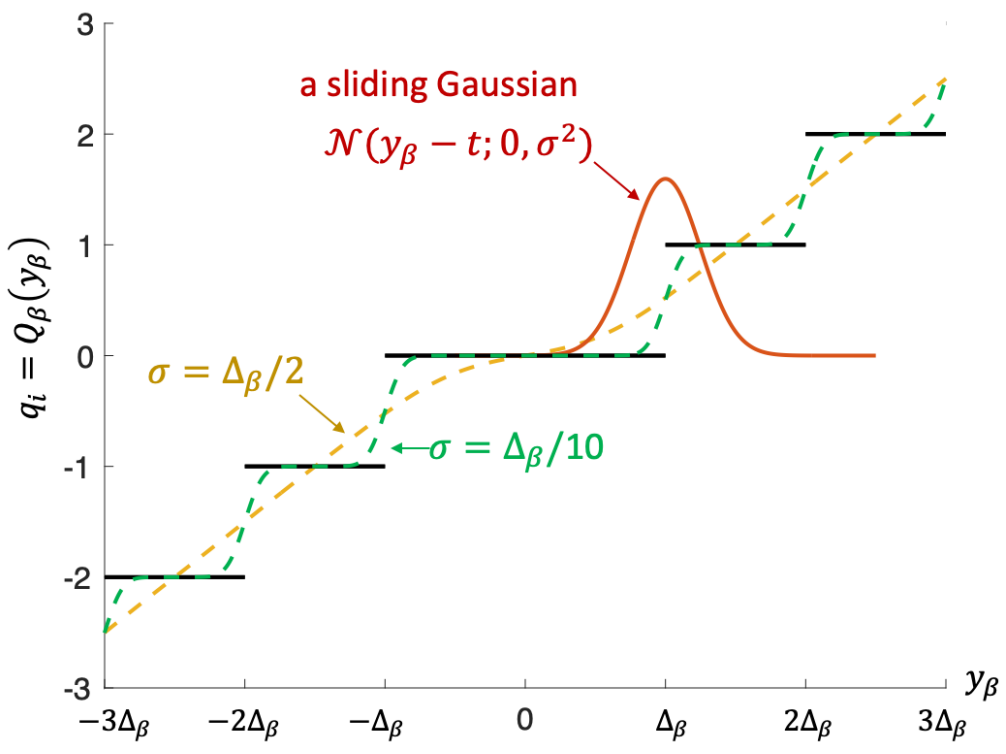

Fig. 17. The proposed backward annealing approach for back-propagation, which interpolates the discontinuous step function $Q_\beta$ (solid black lines) using a sliding Gaussian (red solid curves) to form a differentiable relaxation $\widetilde{Q}_\beta$ (green or yellow dashed curves).

employed in any JPEG 2000 coding framework:

$$q_{i,\beta} = Q_\beta(y_\beta) = \begin{cases} \text{sign}(y_\beta)\left\lfloor \frac{|y_\beta|}{\Delta_\beta} + \xi \right\rfloor, & \frac{|y_\beta|}{\Delta_\beta} + \xi > 0 \\ 0, & \text{otherwise} \end{cases} \quad (4)$$

where $\Delta_\beta$ denotes the quantization step size of subband $B_\beta$ while $\xi$ controls the width of the deadzone. In this paper, $\xi$ is set to be 0, which results in a zero-bin width of $2\Delta_\beta$.

Using Fig. 17 for guidance, we propose to convolve the discontinuous quantization function $Q_\beta$ with a sliding Gaussian function $\mathcal{N}(t;\mu,\sigma^2)|_{\mu=0}$; this convolution relaxes the non-differentiable quantization function $Q_\beta$, producing the continuous relaxation counterpart $\widetilde{Q}_\beta$, which is more suitable for back-propagation. More concretely, we have

$$\widetilde{Q}_\beta(y_\beta) = \int_{-\infty}^{+\infty} Q_\beta(t)\mathcal{N}(y_\beta - t; 0, \sigma^2)dt \quad (5)$$

In practice, we limit the integration to the interval $\pm 3\sigma$, since a normal distribution decays to approximately zero at the endpoints of this interval. With a relatively big $\sigma$, our approach draws a straight line through the quantization step function for the purpose of back-propagation, as exemplified in yellow dashed curve in Fig. 17; this is essentially the concept of the STE. By decreasing the value of $\sigma$ along with the learning rate during the training, the method ensures a smooth transition from STE to soft-to-hard annealing with a controllable “cooling” coefficient $\sigma$; an example is given as the green dashed curve in Fig. 17.

Now we move on to the calculation of the coded length $l_{i,\beta}$ in (2). Conceptually, if we knew the statistical distribution $P_{V_\beta}$, then $l_{i,\beta}$ can be calculated directly using (3). The challenge is that $P_{V_\beta}$ is data-dependent, and depends weakly on the choice of the weights in the high-to-low and the low-to high networks. This weak dependency allows us, in practice, to estimate and update each $P_{V_\beta}$ periodically using a histogram. This histogram, containing the number of occurrence hence the probability of each $q_{i,\beta}$, can then be converted to the coded length $\hat{l}_{i,\beta}$ as an estimate of $l_{i,\beta}$. All the coded lengths $\hat{l}_{i,\beta}$ for all $i \in B_\beta$ eventually form a look-up table $T_\beta$ for each subband $B_\beta$ as depicted in Fig. 16. In terms of the back-propagation, each look-up table $T_\beta$ together with its respective quantizer $Q_\beta$ can be treated like one discontinuous operator, which maps each input $y_\beta$ to its coded length $\hat{l}_{i,\beta}$; this discontinuous operator can be treated exactly the same way as described in Fig. 17.

It is important to highlight the fact that this periodic update of the histograms, therefore the look-up tables $\{T_\beta\}_\beta$, does not introduce instability into the training. Essentially, the training process with periodic update alternates between two steps, each of which reduces the following modified version of the cost function in (2):

$$J'(\phi, \{T_\beta\}_\beta) = \|x - \hat{x}(\phi)\|^2 + \lambda_1 \sum_\beta \sum_{i \in B_\beta} \hat{l}_{i,\beta} + \lambda_2 \sum_d \left\|\widetilde{\mathbf{y}}_{LL,d}(\phi) - \widetilde{\mathbf{y}}^t_{LL,d}\right\|_2^2 \quad (6)$$

where

$$\hat{l}_{i,\beta} = T_\beta(q_{i,\beta}) \quad (7)$$

in which $\phi$ represents the vector of all network weights; we remind the reader that we use the same set of weights for all levels of decomposition.

This modified cost function involves jointly optimizing the look-up tables $\{T_\beta\}_\beta$ and the weights $\phi$, which proceeds in alternating steps. At any given point during training, we have *Step 1* that optimizes the network weights $\phi$, assuming fixed look-up tables $\{T_\beta\}_\beta$. In *Step 2*, we adjust the look-up tables $\{T_\beta\}_\beta$ using histograms, given the set of network weights $\phi$. Each step progressively reduces the same finite bounded objective in (6), therefore the entire system must converge. Although the global objective in (6) that we minimize is not the exactly same as (2), the difference gradually reduces as the periodic update progresses.

## 6 Experimental Results

To explore the merits of our method, we develop a sequence of experiments to test the significance of the aliasing suppression term in (2). In addition, we explore our method with different base wavelet transforms – the LeGall 5/3 and the CDF 9/7 wavelet transforms. To put these results in context, we also compare them with some existing works. Note that the source code of our method, along with all training and testing datasets, are available on GitHub[2].

### 6.1 Experimental Settings

#### *6.1.1 Training Phase*

We employ 5 levels of the DWT decomposition during training, and aim to jointly train only a single pair of high-to-low and low-to-high networks, which can be progressively applied to all levels of decomposition, as well as a wide range of compression ratio. This goal is explicitly chosen, because it is more sensible for practical applications to employ a method which only uses one set of weights for all levels. This is especially important for scalable codecs, where the number of levels received at the decoder may not be the same as the encoder.

Now we begin by discussing the initialization of our training process. As explained in Section 5, there are two alternating update steps during training to minimize the modified objective (6). To start with *Step 1*, we first find the

2. https://github.com/xinyue-li3/hybrid-lifting-structure/

initial look-up tables $\{T_\beta\}_\beta$ using the weights of the high-to-low and the low-to-high networks $\phi$ as trained in the exploration phase in Section 4. Then the network weights $\phi$ can be optimized in *Step 2* given the initial lookup tables $\{T_\beta\}_\beta$, employing the training strategy in Section 5. *Step 1* and *Step 2* alternate periodically (in this paper every 200 epochs), so that (6) gradually converges to (2).

Our initial choices of parameters are based upon the base wavelet transform that we are attempting to improve during the learning process. This base wavelet transform has extra distortion gains $G_{a,\beta} = G_{s,\beta} = 1.0$ as depicted in Fig. 16; we therefore start from this point. In addition, this base wavelet transform also involves different quantization step size $\Delta_\beta$ for each subband $B_\beta$ as seen in (4); we initialize these $\{\Delta_\beta\}_\beta$ in a way which typically results in a compression bit-rate around $1.0$ bpp for training images. This bit-rate $1.0$ bpp also corresponds to a particular rate-distortion slope $\lambda_1$ as seen in (2); therefore, this becomes the starting point of our $\lambda_1$ during training. Subsequently, we then allow $G_{a,\beta}$ and $G_{s,\beta}$ to evolve during training while keeping $\Delta_\beta$ and $\lambda_1$ fixed. If the change in $G_{a,\beta}$ and $G_{s,\beta}$ are not too substantial, then we expect the compression bit-rate to still wind up in the vicinity of 1.0 bpp at the end of training.

In this paper, Keras with TensorFlow backend and the Adam algorithm [32] are employed for training, with 75 image batches comprising 16 patches of size 256 x 256 from the DIV2K image dataset. In total 1200 epochs are used for training in this paper, while periodic update of the look-up tables $\{T_\beta\}_\beta$ occurs every 200 epochs, as mentioned before. Within each 200 epochs, we progressively reduce the controllable "cooling" coefficient $\sigma$ for back-propagation as explained in Section 5. This $\sigma$ is empirically initialized as $\frac{\Delta}{2}$ and decays steadily until the change in the solution is negligible; in our case, this end point yields $\sigma \approx \frac{\Delta}{10}$. The learning rate is empirically set to $0.0001$ and decays exponentially with *decay_steps* $= 20$ and *decay_rate* $= 0.96$. This might not be the optimal training schedule for $\sigma$, but it turns out that other more natural training schedules are hard to realize within the TensorFlow backend.

#### *6.1.2 Testing Phase*

We choose four datasets categorized into three classes during testing, in order to demonstrate the merits of our method in different scenarios. Note that none of these images are used during training.

**Category 1**: All images within this class have highly structured features, i.e. edges are either consistently oriented or significantly distinct from background textures. Two datasets are included in this class: a) Tecnick Sampling Dataset[3], from which 20 images are chosen with size 480 x 480; b) DIV2K Dataset[4], from which 30 images are chosen with size 1024 x 2048. We name these two dataset as *Tecnick-Cat1* and *DIV2K-Cat1*, respectively.

**Category 2**: All images in this category come with reasonably clear edges, while background textures are more complicated than those in Category 1. The dataset employed in this class is DIV2K dataset, from which another 70 images of size 1024 x 2048 are chosen; it is denoted by *DIV2K-Cat2*.

**Category 3**: All images in this category are considered to be "hard-to-code", with one or more following properties: nearly no clear orientations; majority of the image is excessively blurred; and/or most orientations are horizontal or vertical, which are well handled by the wavelet transform. The dataset employed is Challenges on Learned Image Compression 2019 test set[5], from which 15 images of size 1024 x 2048 are chosen; this is denoted as *CLIC2019-Cat3*.

Moreover, the Kodak dataset[6], which is commonly used as the benchmark for image compression, is also tested here to serve two purposes: 1) to demonstrate the effectiveness of our method in an entire dataset, which is not explicitly chosen nor altered; 2) to put our results in context with other existing works, even if the source codes are unavailable or hard to reproduce the inferences given their codes.

It is worthwhile pointing out that all images within all datasets are converted to grayscale before any training or testing. The reason for this is to avoid confusing our spatial transforms with color dependent questions, such as the optimal choice of color transform and the dependence of the wavelet transform on different color components.

### 6.2 Methods Explored

We first explore the following variations of our method:

- the effectiveness of replacing the adaptive operator $\mathcal{T}^A_{L2H}$ with the linear $\mathcal{T}^W_{L2H}$ in the hybrid architecture; As suggested in Section 3, $\mathcal{T}^W_{L2H}$ might be sufficient.
- three variations of the aliasing constraint parameter in (2): $\lambda_2 = 0$, $\lambda_2 = 1$ and $\lambda_2$ decreasing gradually from 1 to 0 throughout training; although $\lambda_2$ is not directly coupled with coding efficiency, it is driven by visual considerations as explained in Section 5.
- two different base wavelet transforms: the LeGall 5/3 [39] and the CDF 9/7 bi-orthogonal wavelet transforms [40], as they come with different levels of complexity and spatial supports.

To put these results in context, they are also compared with some other existing works from the following categories: i) non-learning based compression standards; ii) learning-based, wavelet-like lossy image compression frameworks; and iii) variants of end-to-end optimized, non-wavelet-like lossy image compression with neural networks.

### 6.3 Evaluation metrics

We consider evaluating the performance of all methods presented within this section both quantitatively and qualitatively. In terms of quantitative measurements, three widely used metrics are employed – Peak Signal-to-Noise Ratio (PSNR), Structural Similarity (SSIM), Multi-Scale Structural Similarity (MS-SSIM). All these metrics are measured and averaged for each dataset, from which Bjøntegaard (BD) rate savings (in %) are obtained.

With regard to qualitative assessment, we provide examples for both the "cleaned" LL bands at different scales and the full reconstructed images in this paper. We remind the readers that the quality of the "cleaned" LL bands is dependent on $\lambda_2$ in (2), which controls the amount of aliasing as explained in Section 5.

3. https://testimages.org/
4. https://data.vision.ee.ethz.ch/cvl/DIV2K/
5. http://clic.compression.cc/2019/challenge/
6. http://www.cs.albany.edu/ xypan/research/snr/Kodak.html

TABLE 2
Comparison between the adaptive operator $\mathcal{T}^A_{L2H}$ and the linear operator $\mathcal{T}^W_{L2H}$. The table shows BD-rate improvements for PSNR, SSIM and MS-SSIM metrics over the LeGall 5/3 wavelet transform for each dataset. Results are obtained with bit-rates between $0.1$bpp to $1.0$bpp.

| | | BD-rate for PSNR | | BD-rate for SSIM | | BD-rate for MS-SSIM | |
|---|---|---|---|---|---|---|---|
| | | $\mathcal{T}^A_{H2L}$ + $\mathcal{T}^A_{L2H}$ | $\mathcal{T}^A_{H2L}$ + $\mathcal{T}^W_{L2H}$ | $\mathcal{T}^A_{H2L}$ + $\mathcal{T}^A_{L2H}$ | $\mathcal{T}^A_{H2L}$ + $\mathcal{T}^W_{L2H}$ | $\mathcal{T}^A_{H2L}$ + $\mathcal{T}^A_{L2H}$ | $\mathcal{T}^A_{H2L}$ + $\mathcal{T}^W_{L2H}$ |
| LeGall 5/3 | Tecknick-Cat1 | $-17.4\%$ | $-8.2\%$ | $-15.5\%$ | $-5.8\%$ | $-13.6\%$ | $-4.7\%$ |
| | DIV2K-Cat1 | $-14.4\%$ | $-7.3\%$ | $-13.8\%$ | $-5.0\%$ | $-13.1\%$ | $-4.5\%$ |
| | DIV2K-Cat2 | $-12.5\%$ | $-6.0\%$ | $-12.8\%$ | $-5.6\%$ | $-12.8\%$ | $-5.9\%$ |
| | CLIC2019-Cat3 | $-7.3\%$ | $-3.4\%$ | $-7.5\%$ | $-2.7\%$ | $-8.9\%$ | $-4.0\%$ |

## 6.4 Results and Discussions

### 6.4.1 *Significance of the adaptive operator*

We first empirically study the value of employing an adaptive low-to-high operator $\mathcal{T}^A_{L2H}$ rather than the linear operator $\mathcal{T}^W_{L2H}$ in the hybrid architecture. As explained in Section 3, $\mathcal{T}^W_{L2H}$ is conceptually sufficient to suppress redundancy within the detail bands, only if $\mathcal{T}^A_{H2L}$ is completely successful in cleaning all aliasing from the LL band; however, we do not expect it to be sufficient in practice.

To study this, the adaptive high-to-low and low-to-high networks ($\mathcal{T}^A_{H2L}$ and $\mathcal{T}^A_{L2H}$) are set as seen in Section 4.3, while the operator $\mathcal{T}^W_{L2H}$ is simply a linear filter that has the same region of support as $\mathcal{T}^A_{L2H}$. $\mathcal{T}^A_{H2L}$ is jointly trained with either $\mathcal{T}^A_{L2H}$ or $\mathcal{T}^W_{L2H}$ to improve the LeGall 5/3 wavelet transform, targeting the standard rate-distortion objective for MSE by setting $\lambda_2 = 0$ in (2). Further studies on different $\lambda_2$ and base wavelet transforms are provided shortly.

The BD rate savings (in %) for average PSNR, SSIM and MS-SSIM over the range of bit-rates from $0.1$bpp to $1.0$bpp across all four datasets are provided in Table. 2; the complete rate-distortion curves can be found in supplementary materials. It can be observed consistently that both $\mathcal{T}^A_{L2H}$ and $\mathcal{T}^W_{L2H}$ are capable of improving coding efficiency of the conventional LeGall 5/3 wavelet transform, regardless the amount of distinct edges presented in images, so long as the adaptive high-to-low operator $\mathcal{T}^A_{H2L}$ is employed. More importantly, $\mathcal{T}^A_{L2H}$ performs significantly better than $\mathcal{T}^W_{L2H}$ across all datasets, achieving up to 17.4% average BD rate saving over the LeGall 5/3 wavelet transform, while $\mathcal{T}^W_{L2H}$ only reaches up to 8.2% average BD bit-rate saving.

This observation aligns with the underlying theory presented in Section 3; by introducing an adaptive low-to-high operator, the hybrid approach can maintain the benefits of coding efficiency even if $\mathcal{T}^A_{H2L}$ is unable to fully clean aliasing from the low-pass band.

### 6.4.2 *Role of the aliasing constraint term*

We now examine the role that the aliasing constraint term, $\lambda_2 \sum_d \left\|\tilde{\mathbf{y}}_{LL,d}(\boldsymbol{\phi}) - \tilde{\mathbf{y}}^t_{LL,d}\right\|_2^2$, plays in our training objective function seen in (2). Specifically, we explore three settings of the aliasing constraint parameter: $\lambda_2 = 0$, $\lambda_2 = 1$ and $\lambda_2$ decreasing from 1 to 0 during training; details have been given under (2) in Section 5. The operators employed here are the adaptive operators $\mathcal{T}^A_{H2L}$ and $\mathcal{T}^A_{L2H}$ as depicted in Section 4.3, whose benefit has been verified. The two operators are jointly trained to improve the LeGall 5/3 wavelet transform at this stage; extension to larger wavelet transforms will be given shortly.

Since $\lambda_2$ explicitly conditions the visual quality of the "cleaned" LL band at each level of decomposition, we now add this additional qualitative assessment into consideration when evaluating the performance of all methods in this section. Using the examples in Fig. 18 as guidance, we can observe that forcing aliasing suppression, i.e. $\lambda_2 = 1$ during training, does indeed ensure higher visual quality of the "cleaned" LL bands across multiple levels of decomposition. Moreover, we observe that employing $\lambda_2 = 1$ or annealing $\lambda_2$ from 1 to 0 produces substantially similar results.

In addition, we provide the BD rate saving (in %) under average PSNR, SSIM and MS-SSIM over the range of bit-rates from $0.1$bpp to $1.0$bpp for different $\lambda_2$ settings across all datasets in Table. 3; the complete rate-distortion curves can be found in supplementary materials. We can first see that, although the two networks $\mathcal{T}^A_{H2L}$ and $\mathcal{T}^A_{L2H}$ are jointly optimized to minimize MSE during training, they also work surprisingly well under the SSIM and the MS-SSIM metrics for all datasets. Not surprisingly, we observe the highest BD bit-rate saving when $\lambda_1 = 0$. However, the loss in coding efficiency associated with $\lambda_2 = 1$ and annealing $\lambda_2$ is not significant, in exchange for clear benefits obtained in visual quality at reduced resolutions.

It is also worthwhile to point out that the performance of our method does vary for different types of images. For Tecnick-Cat1 and DIV2K-Cat1 datasets that have consistent orientations or distinct edges, our method has the highest performance, achieving 17.4% average BD bit-rate saving over the LeGall 5/3 wavelet transform over the range of bit-rate from $0.1$ bpp to $1.0$ bpp. For DIV2K-Cat2 dataset with richer textures, the proposed method also manages to reach 12.8% average BD bit-rate saving over the range of bit-rate from $0.1$ bpp to $1.0$ bpp. Surprisingly, for images in CLIC2019-Cat3 dataset, which comes with hardly any clear orientations and edges, our method is still capable of achieving 8.94% average BD bit-rate saving. These observations again align with our underlying assumption in Section 2, that orientation is the key factor to reduce redundancy (notably aliasing) from the wavelet subbands to improve coding efficiency.

In the end, we also inspect the quality of reconstructed images at different bit-rates from various images. Some examples are given in Fig. 20. We see that the proposed method produces significantly better reconstructed images, with less ringing around edges and more recovered textures than the conventional LeGall 5/3 wavelet transform at similar bit-rates.

TABLE 3
Comparison between different aliasing constraint parameters $\lambda_2$ in (2) during training. The table shows BD-rate improvements for PSNR, SSIM and MS-SSIM metrics over the LeGall 5/3 and the CDF 9/7 wavelet transform. Results are obtained with bit-rates between $0.1$bpp to $1.0$bpp.

| | | BD-rate for PSNR | | | BD-rate for SSIM | | | BD-rate for MS-SSIM | | |
|---|---|---|---|---|---|---|---|---|---|---|
| | | $\lambda_2 = 0$ | $\lambda_2 = 1$ | anneal $\lambda_2$ | $\lambda_2 = 0$ | $\lambda_2 = 1$ | anneal $\lambda_2$ | $\lambda_2 = 0$ | $\lambda_2 = 1$ | anneal $\lambda_2$ |
| LeGall 5/3 | Tecknick-Cat1 | $-17.4\%$ | $-13.8\%$ | $-13.6\%$ | $-15.5\%$ | $-12.5\%$ | $-12.3\%$ | $-13.6\%$ | $-10.3\%$ | $-10.1\%$ |
| | DIV2K-Cat1 | $-14.4\%$ | $-10.6\%$ | $-10.6\%$ | $-13.8\%$ | $-10.8\%$ | $-10.9\%$ | $-13.1\%$ | $-8.9\%$ | $-9.0\%$ |
| | DIV2K-Cat2 | $-12.5\%$ | $-9.8\%$ | $-9.8\%$ | $-12.8\%$ | $-10.6\%$ | $-10.5\%$ | $-12.8\%$ | $-9.8\%$ | $-9.8\%$ |
| | CLIC2019-Cat3 | $-7.3\%$ | $-5.8\%$ | $-5.7\%$ | $-7.5\%$ | $-5.8\%$ | $-5.9\%$ | $-8.9\%$ | $-5.7\%$ | $-5.8\%$ |
| CDF 9/7 | Tecknick-Cat1 | $-11.4\%$ | $-9.8\%$ | $-9.7\%$ | $-11.5\%$ | $-10.6\%$ | $-10.5\%$ | $-9.0\%$ | $-7.9\%$ | $-7.9\%$ |
| | DIV2K-Cat1 | $-9.7\%$ | $-6.3\%$ | $-6.3\%$ | $-10.1\%$ | $-8.4\%$ | $-8.3\%$ | $-7.6\%$ | $-5.3\%$ | $-5.3\%$ |
| | DIV2K-Cat2 | $-7.6\%$ | $-6.0\%$ | $-6.0\%$ | $-7.5\%$ | $-6.7\%$ | $-6.7\%$ | $-5.9\%$ | $-4.9\%$ | $-5.0\%$ |
| | CLIC2019-Cat3 | $-4.2\%$ | $-3.4\%$ | $-3.3\%$ | $-4.1\%$ | $-3.5\%$ | $-3.4\%$ | $-1.7\%$ | $-0.8\%$ | $-0.7\%$ |

#### 6.4.3 *Extension to larger wavelet transform*

Although the structures of our neural networks have been developed in the first instance for the LeGall 5/3 wavelet transform, exactly the same network structures turn out to be also effective with the CDF 9/7 wavelet transform.

Specifically, we jointly train the two adaptive operators $\mathcal{T}^A_{H2L}$ and $\mathcal{T}^A_{L2H}$ to improve the CDF 9/7 wavelet transform, with all three settings of the aliasing constraint parameter $\lambda_2$. Similar as before, we evaluate the performance of the proposed method both quantitatively and qualitatively.

Examples of the visual quality of the "cleaned" LL bands are shown in Fig. 19. Although the CDF 9/7 wavelet transform already produces less aliased LL bands, our method still manages to reduce the remaining aliasing and produce more visually appealing LL bands. From Fig. 21 we also see that the visual quality of the reconstructed images from our method is significantly better than that from the CDF 9/7 wavelet transform, with much less ringing around edges; more results can be found in supplementary materials.

Regarding the quantitative performance, our method can still achieve up to $11.5\%$ average BD bit-saving over the range of bit-rate from $0.1$ bpp to $1.0$ bpp, as seen in Table. 3. The story is fairly consistent for the SSIM and MS-SSIM metrics as well; the complete rate-distortion curves can be found in supplementary materials. All these results align with our previous conclusions with the LeGall 5/3 wavelet transform.

#### 6.4.4 *Comparison with existing works*

To put our method in context, we compare the variations of the proposed approach with some existing works. These well-known works are: i) the JPEG2000 (with the LeGall 5/3 and the CDF 9/7 wavelets) and the WebP compression standards, which do not involve any machine learning; ii) iWave [12] and Dardouri [14], which are machine-learning optimized lifting schemes for wavelet-like lossy image compression; iii) Theis [20], Toderici [22] and Johnston [23], which are variants of the end-to-end optimized, non-wavelet-like learned lossy image compression systems.

For the sake of this comparison, we prefer to avoid including methods that employ a dedicated post-processing step on reconstructed data to reduce artifacts like in [11] or very sophisticated context modeling for entropy coding, such as [18] [41]. Moreover, to the best of our knowledge, all the end-to-end optimized image compression systems, including Theis [20], Toderici [22] and Johnston [23], lack important attributes of our method, such as resolution scalability, quality scalability and accessibility to region-of-interest. The resolution scalability feature, however, is found in iWave [12] and Dardouri [14], making comparisons with these methods particularly interesting. At the same time, these wavelet-like methods do not explicitly consider the visual quality of the LL bands at different resolutions, and they do not propose an end-to-end training strategy to directly optimize their methods for rate-distortion objective.

Fig. 22 provides the average PSNR and MS-SSIM results using the commonly tested Kodak dataset. The proposed method appears to be very competitive with other existing methods. Interestingly, the PSNR performance of iWave [12] is very close to our method, further confirming that wavelet-like compression schemes can be competitive with end-to-end optimized non-wavelet-like methods. For the other wavelet-like method Dardouri [14], we are unable to execute their inference procedure available to us. However, we observe from [14] that they are unable to present competitive PSNR and MS-SSIM results with respective to JPEG2000.

#### 6.4.5 *Computational complexity*

Finally, we evaluate computational complexity as well as region of support associated with our method, in comparison with other existing works. From Table 4 we can see, our method comes with the fewest number of parameters and relatively small region of support.

TABLE 4
Comparisons of computational complexity and region of support

| | Num of Parameters | Region of Support |
|---|---|---|
| Our method | $33K$ | 37 x 37 |
| JPEG2000 [2] | - | 9 x 7 or 5 x 3 |
| WebP | - | - |
| iWave [12] | $97K$ | 25 x 25 |
| Dardouri [14] | $167K$ | 253 x 125 |
| Toderici [22] | $5.4M$ | 250 x 250 |
| Theis [20] | $3.3M$ | 43 x 43 |
| Johnston [23] | $9.9M$ | 300 x 300 |

## 7 Conclusion and Discussions

In this paper, we propose two networks, the high-to-low and low-to-high networks, as additional lifting steps to augment the conventional wavelet transforms, improving coding efficiency and visual quality of LL bands across multiple levels of decomposition. The high-to-low network serves to clean aliasing and perhaps other redundancies from the low-pass band produced at each successive level of the decomposition, while the low-to-high network aims to further reduce redundancy amongst the detail bands.

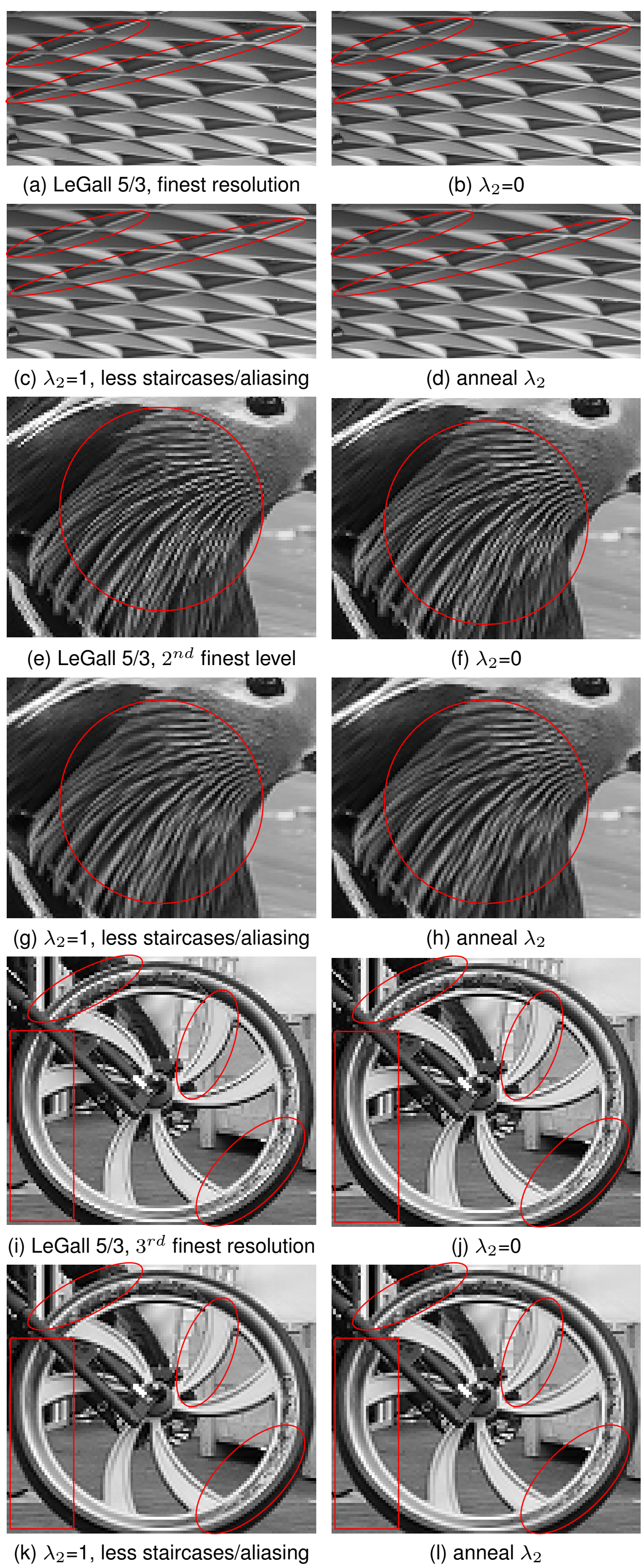

(a) LeGall 5/3, finest resolution (b) $\lambda_2$=0

(c) $\lambda_2$=1, less staircases/aliasing (d) anneal $\lambda_2$

(e) LeGall 5/3, $2^{nd}$ finest level (f) $\lambda_2$=0

(g) $\lambda_2$=1, less staircases/aliasing (h) anneal $\lambda_2$

(i) LeGall 5/3, $3^{rd}$ finest resolution (j) $\lambda_2$=0

(k) $\lambda_2$=1, less staircases/aliasing (l) anneal $\lambda_2$

Fig. 18. Visual quality of the “cleaned” LL bands at different scales from various images, obtained using different $\lambda_2$ strategies during training, and optimized for the LeGall 5/3 wavelet transform. Note that by reducing aliasing, we produce smoother edges with less “staircase” artifacts.

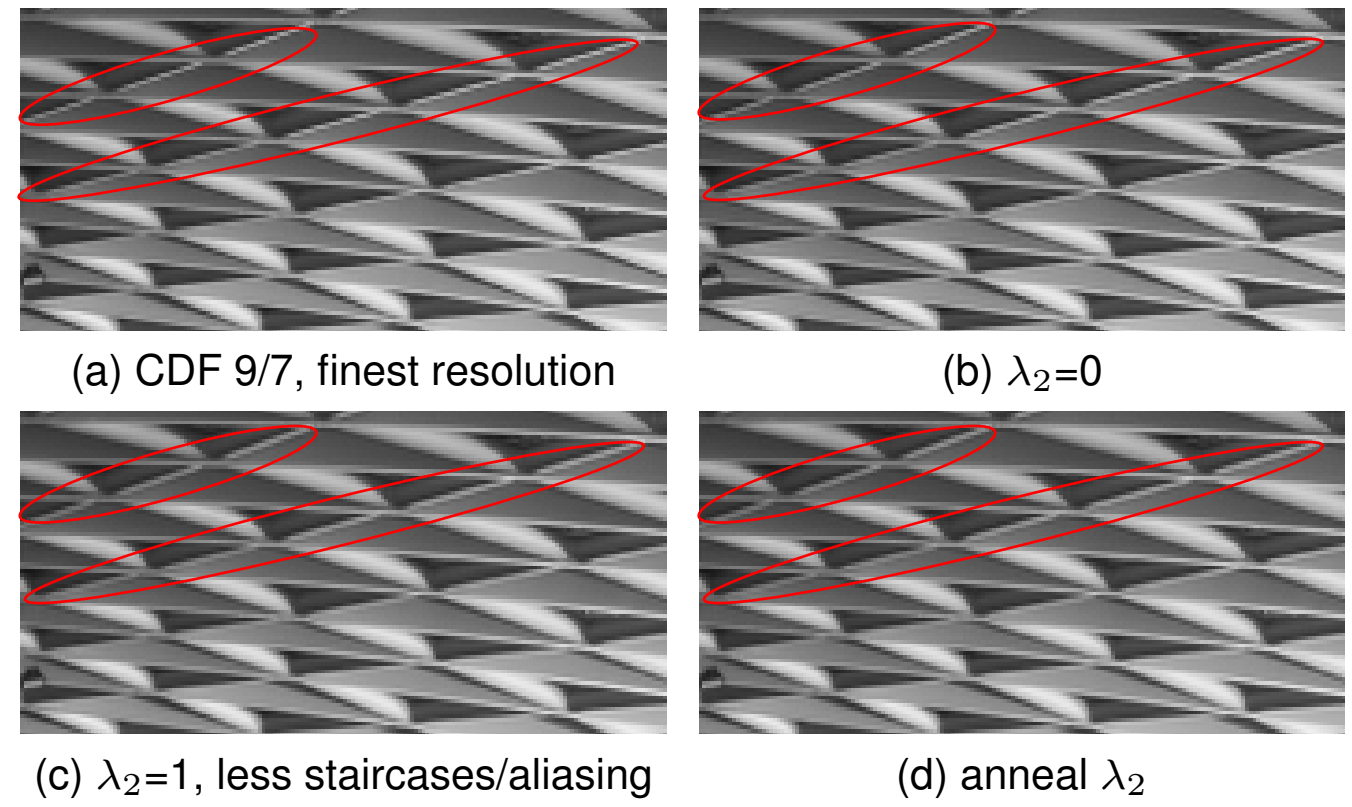

(a) CDF 9/7, finest resolution (b) $\lambda_2$=0

(c) $\lambda_2$=1, less staircases/aliasing (d) anneal $\lambda_2$

Fig. 19. Visual quality of the “cleaned” LL bands at the finest resolution, obtained using different $\lambda_2$ strategies during training, and optimized for the CDF 9/7 wavelet transform. Note that by reducing aliasing, we produce smoother edges with less visible “staircase” artifacts.

Our proposed approach is inspired and guided by a specific theoretical argument related to the opportunity presented by geometric flow and connected to super resolution. Specifically, this argument reveals the role that geometric flow can play in untangling redundant information from the low- and the high-pass subbands. Following the ablation study of different network structures, we eventually find that the best investigated solution does indeed involve banks of optimized linear filters controlled dynamically by an opacity network, as suggested by the underlying theory.

More importantly, the networks driven by our hypothesis are compact and with limited non-linearities, allowing high coding efficiency and scalability over a wide range of bit-rates and multiple resolutions. This means that all coded *wavelet* coefficients have a relatively small region of influence in the image domain, and there is no need to learn and store separate network weights for each decomposition level. In addition, since the structure involves a collection of purely linear filters, our method comes with a fairly low computational complexity.

Apart from the networks themselves, we also propose a simple yet appealing relaxation approach to manage discontinuities in quantization and cost functions during training, so as to jointly train the proposed networks in an end-to-end fashion. By selectively including the aliasing suppression term in the training objective, we demonstrate that the proposed method achieves up to $17.4\%$ average BD rate saving over the conventional wavelet transforms in a wide range of bit-rates. Moreover, our method also manages to reduce aliasing at intermediate resolutions, producing a more visually appealing multi-scale wavelet representation. Overall, the coding efficiency of the proposed scheme appears to be very competitive with other related works,

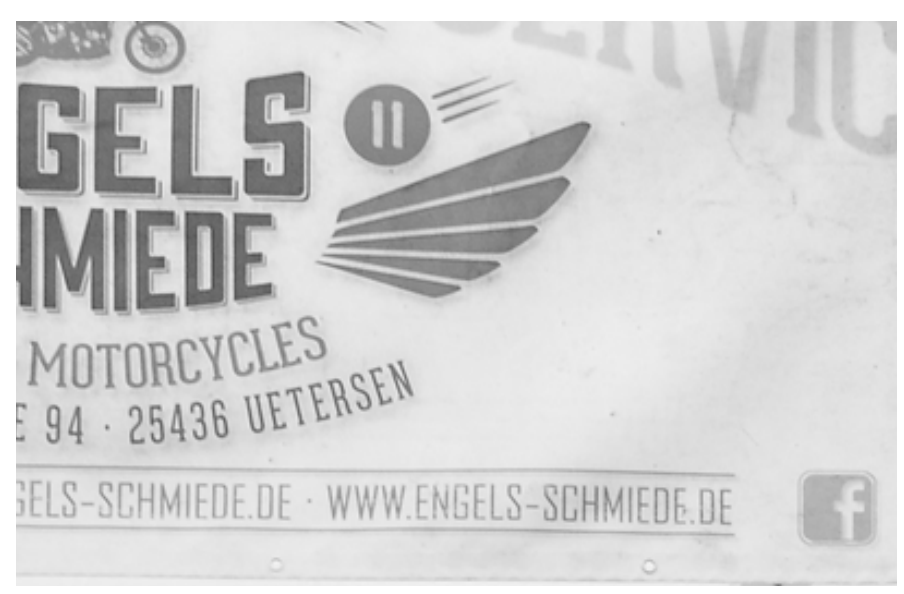

(a) Original image, cropped from image 28 of DIV2K-Cat2 dataset

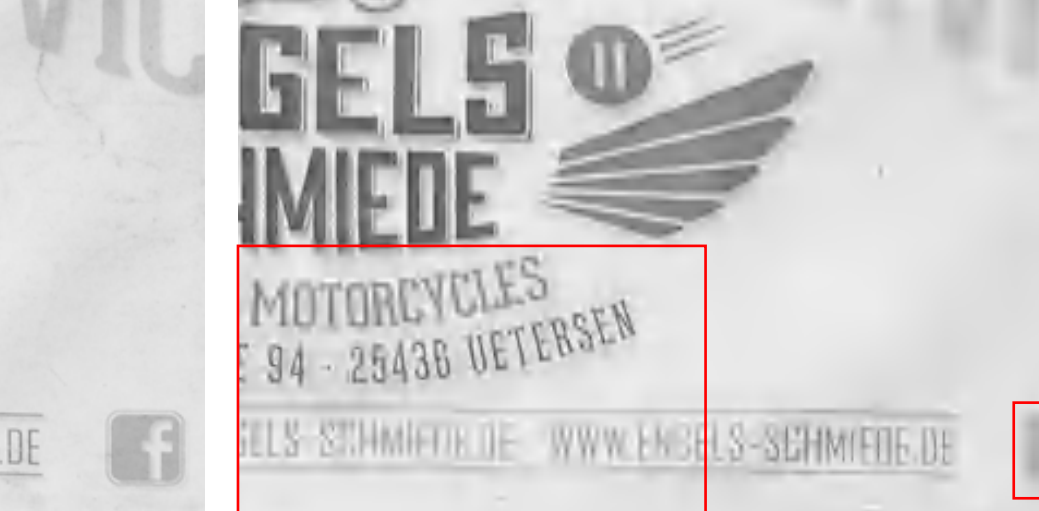

(b) LeGall 5/3 at 0.199bpp, PSNR=29.53dB, SSIM=0.841, MS-SSIM=0.9585

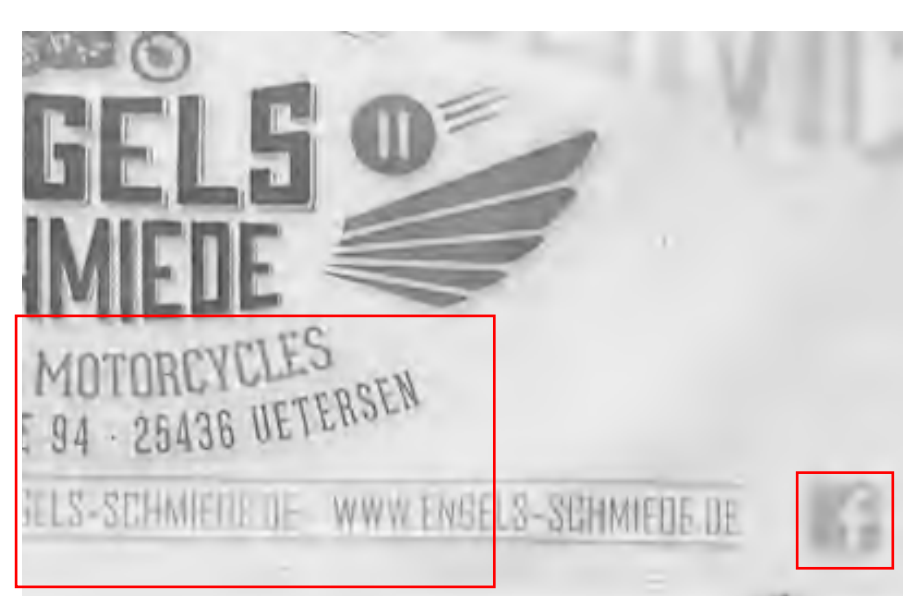

(c) Proposed,$\lambda_2$=0,0.197bpp,PSNR=30.09dB, SSIM=0.854,MS-SSIM=0.9631

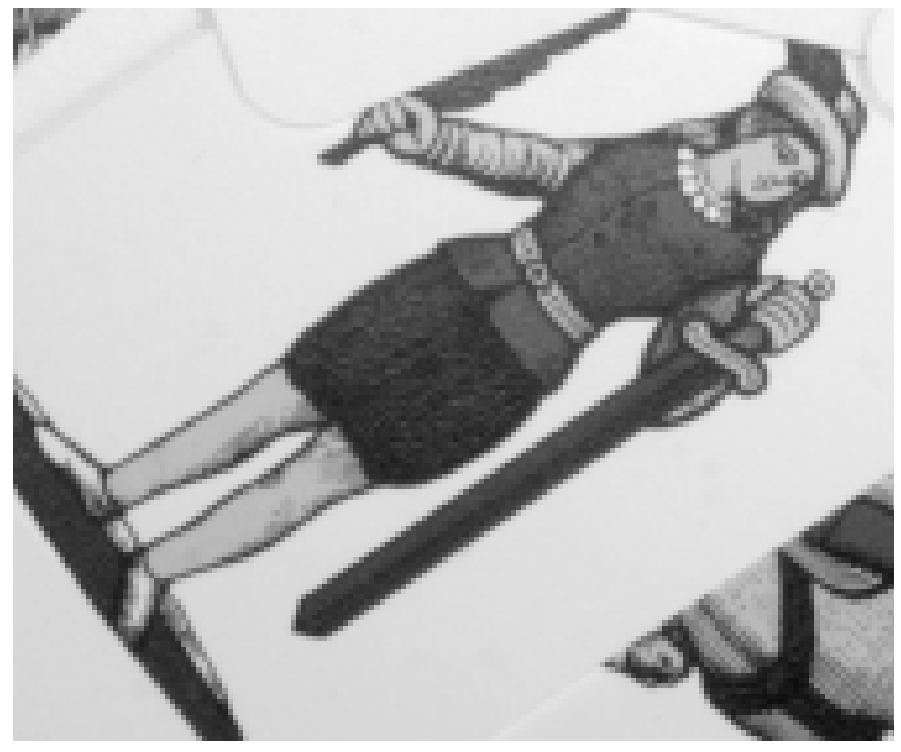

(d) Original image, cropped from image 9 of Tecnick-Cat1 dataset

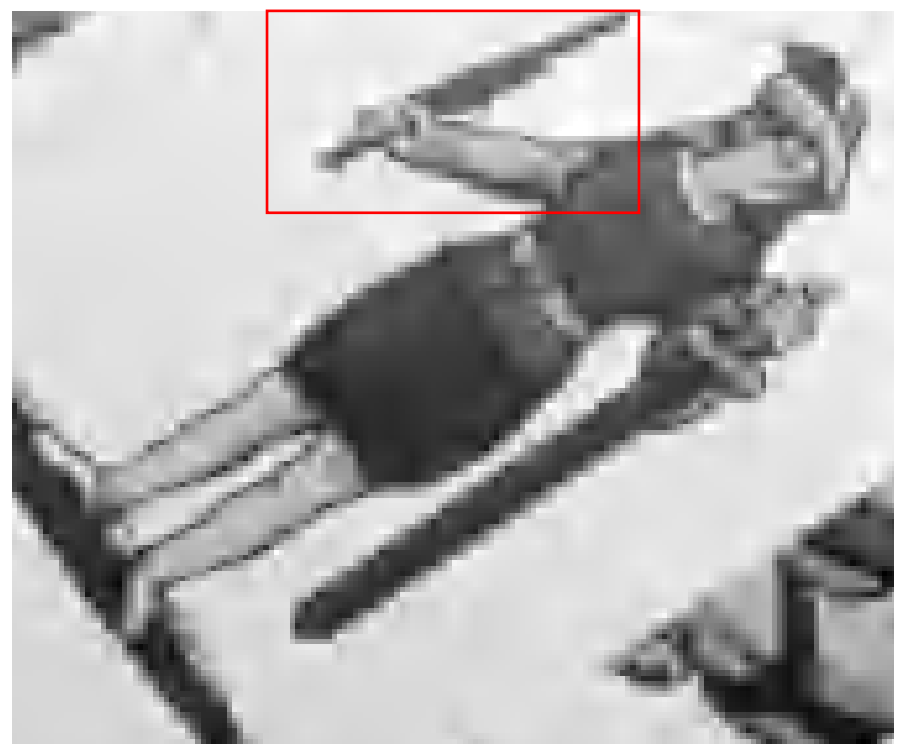

(e) LeGall 5/3 at 0.298bpp, PSNR=26.97dB, SSIM=0.862, MS-SSIM=0.9694

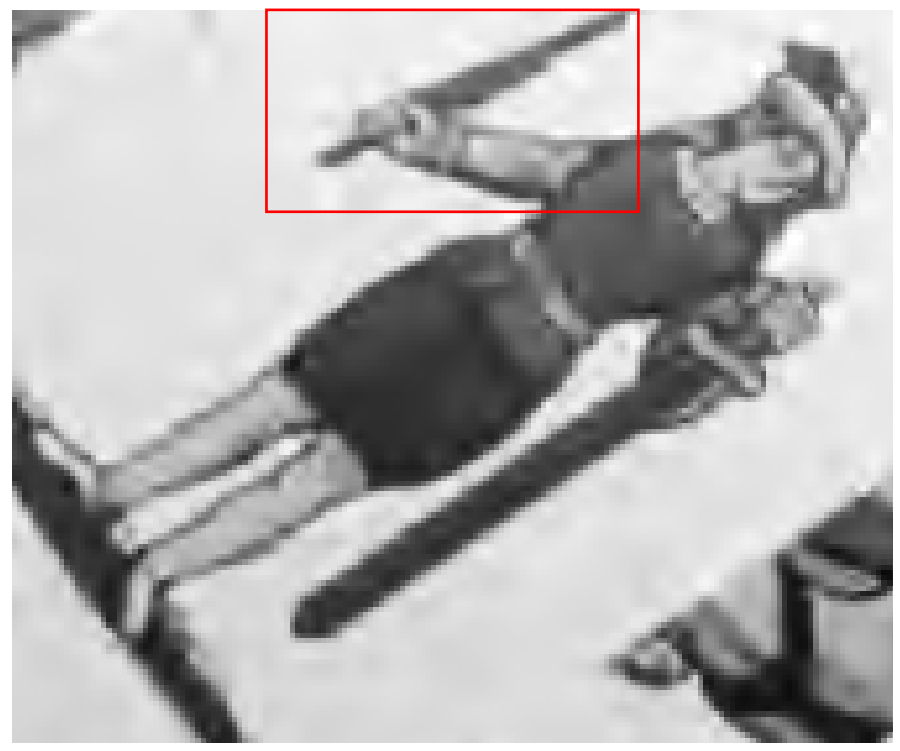

(f) Proposed,$\lambda_2$=0,0.295bpp,PSNR=28.19dB, SSIM=0.875,MS-SSIM=0.9718

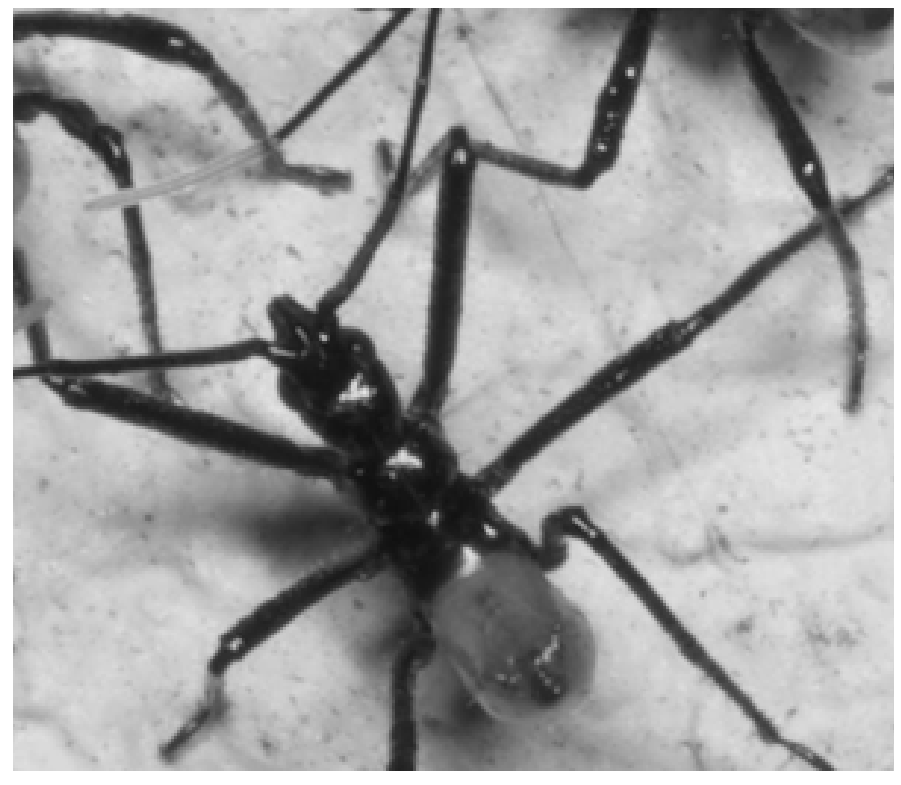

(g) Original image, cropped from image 28 of DIV2K-Cat1 dataset

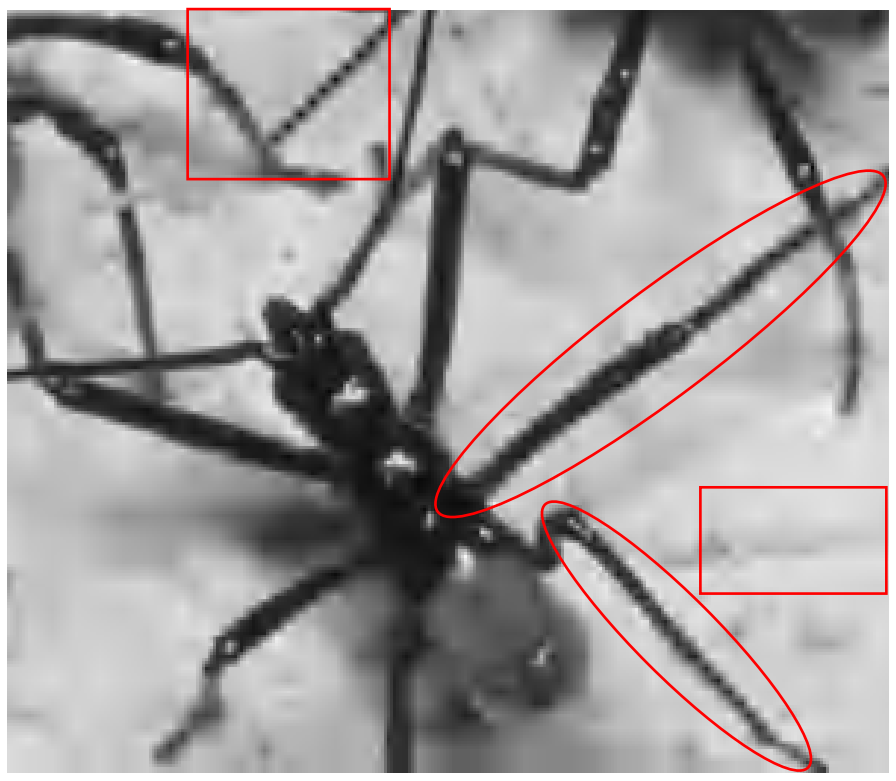

(h) LeGall 5/3 at 0.197bpp, PSNR=29.27dB, SSIM=0.831, MS-SSIM=0.9526

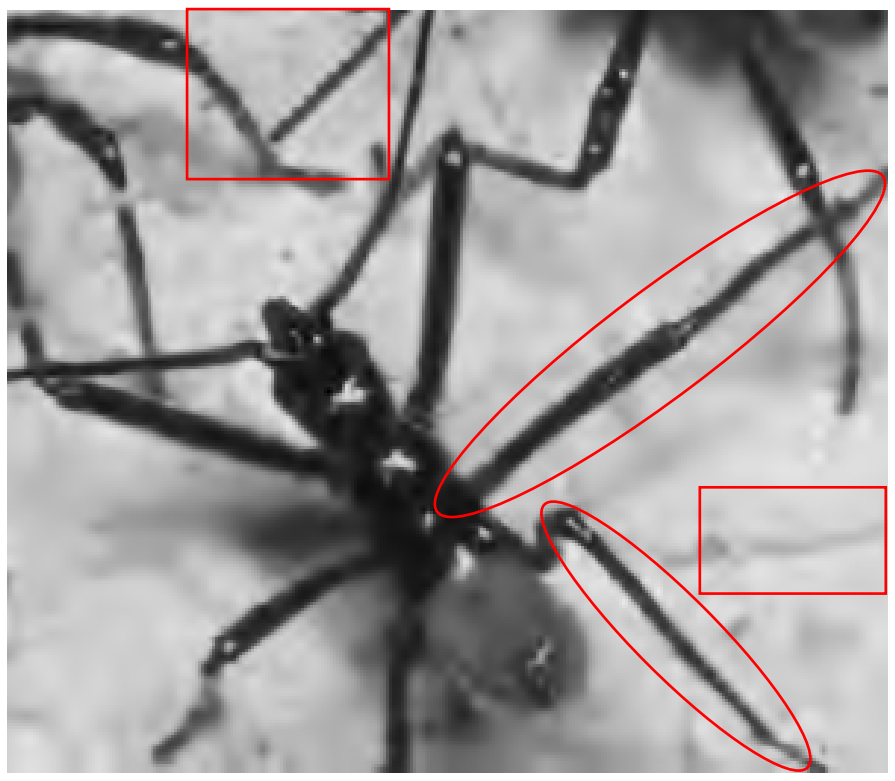

(i) Proposed,$\lambda_2$=0,0.196bpp,PSNR=30.39dB, SSIM=0.850, MS-SSIM=0.9592

Fig. 20. Examples of reconstructed images. The proposed method is trained to improve the conventional LeGall 5/3 wavelet transform with $\lambda_2 = 0$.

meanwhile offering other desirable features like resolution and quality scalability and region-of-interest accessibility.

There are many possibilities to improve upon our current work. Our current usage of machine learning is modest with relatively small number of parameters and layers. There is, of course, the opportunity to extend the number of lifting steps and the nature of the network structures inside these lifting steps. Moreover, it it possible to incorporate additional features, such as learned color transforms, collections of rich context models and learned post-processing strategies. All of these additional features add complexities and may have an adverse impact on the scalability of the scheme, therefore may not be valuable for practical applications. Amongst all the things that could be done, some are worthier investigations than others. One important point for investigation is selective replacement or augmentation of the lifting steps within the base wavelet transforms with learning techniques. Another is to investigate the potential of the network structures similar to the one presented in this paper, but with many more channels (currently only four in this paper).

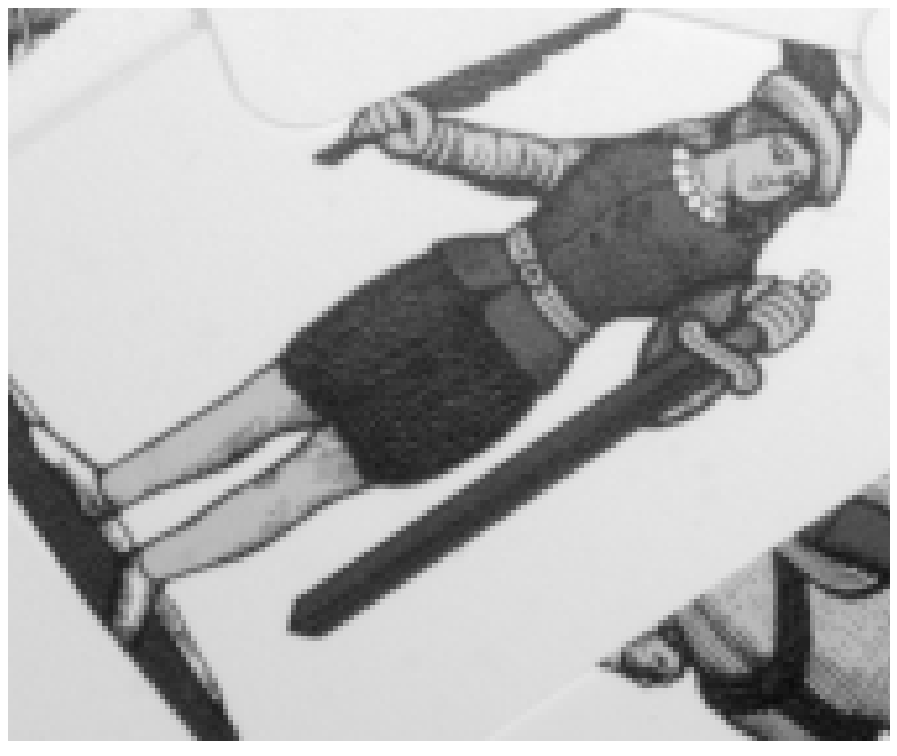

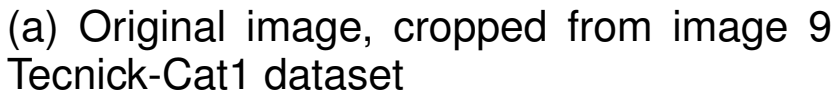
(a) Original image, cropped from image 9 of Tecnick-Cat1 dataset

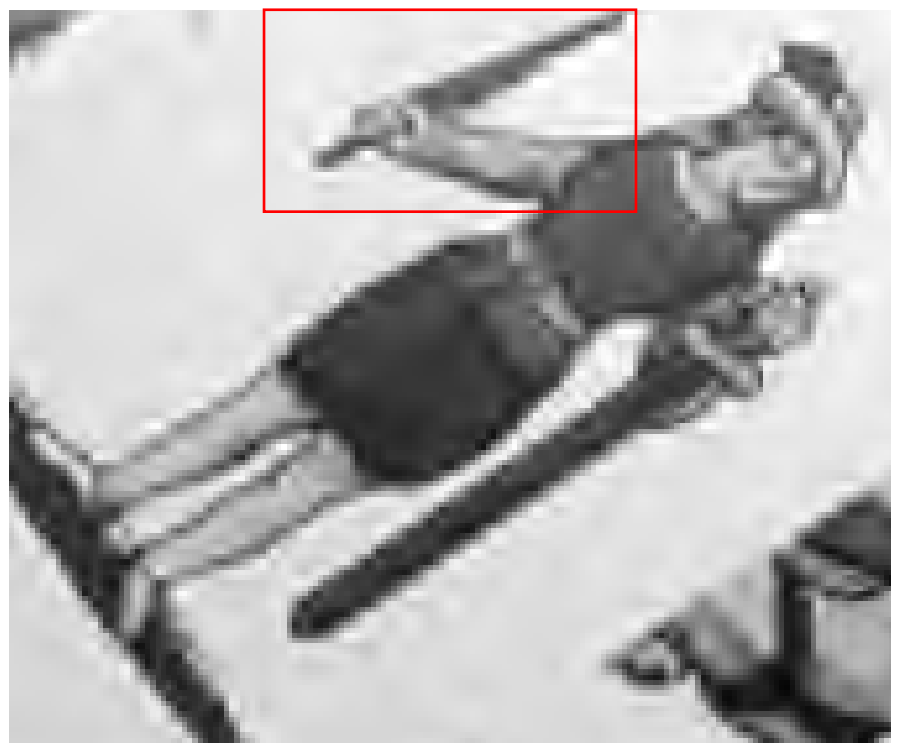
(b) CDF 9/7 at $0.295$bpp, PSNR=$27.36$dB, SSIM=$0.854$, MS-SSIM=$0.9681$

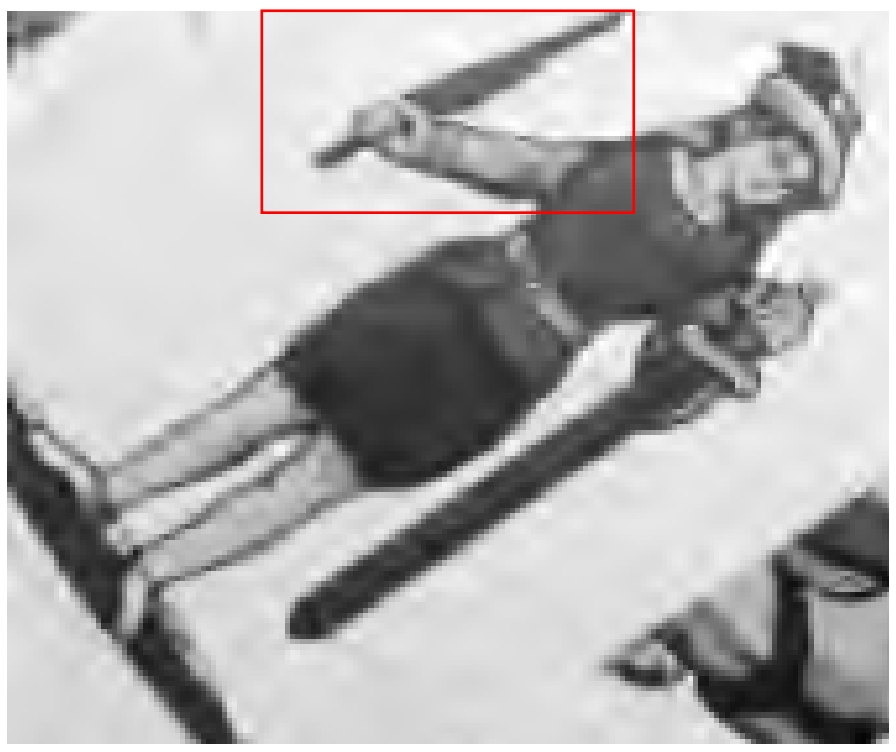
(c) Proposed,$\lambda_2$=0,$0.295$bpp,PSNR=$28.41$dB, SSIM=$0.870$,MS-SSIM=$0.9711$

Fig. 21. Examples of reconstructed images. The proposed method is trained to improve the conventional CDF 9/7 wavelet transform with $\lambda_2 = 0$.

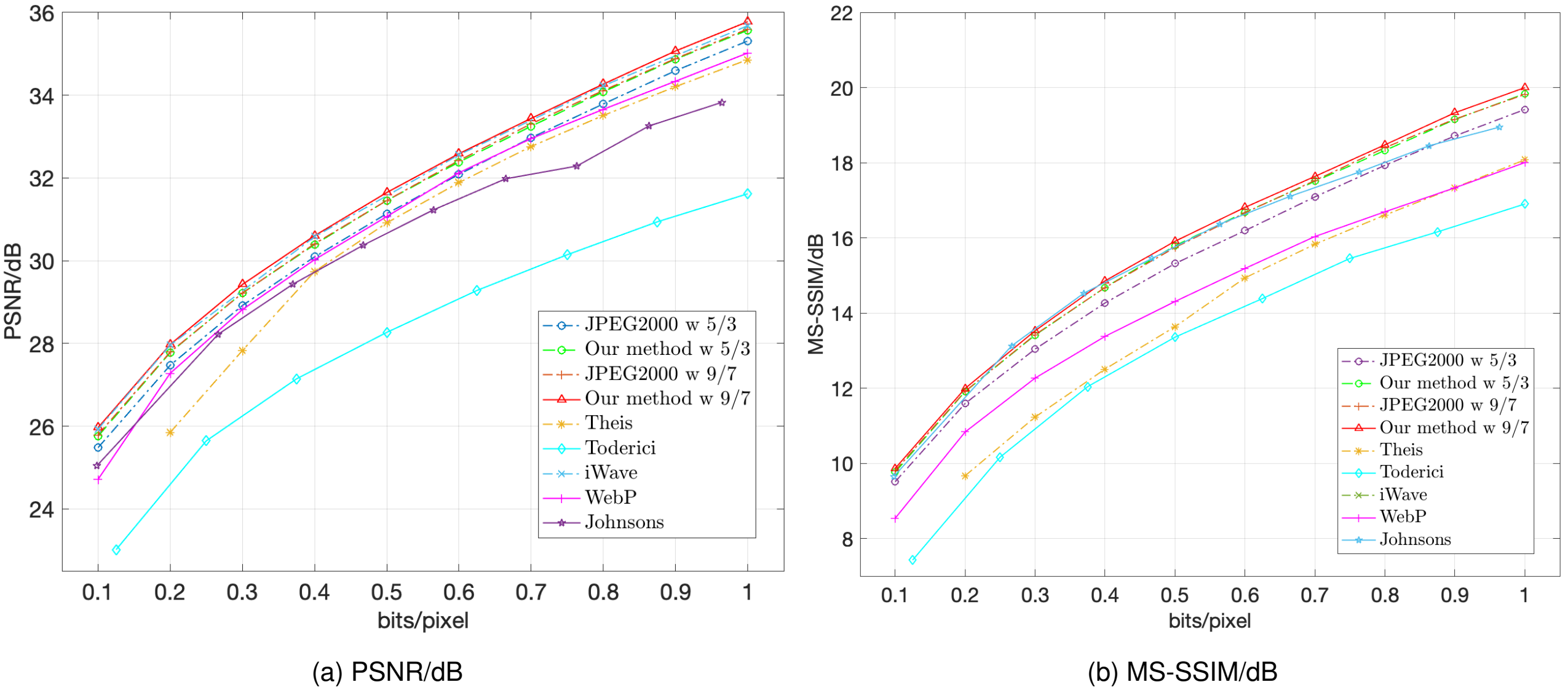

(a) PSNR/dB

(b) MS-SSIM/dB

Fig. 22. Comparisons of the average PSNR and MS-SSIM improvements between our methods and other existing works for the Kodak Dataset. MS-SSIM are calculated in dB as: $-10\log_{10}(1-$MS-SSIM$)$. The results of other works are taken from the original papers without any reproduction.

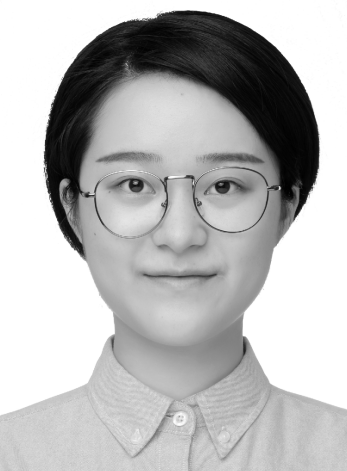

**Xinyue Li** Xinyue Li received the M.Eng degree (with Excellence) in electrical engineering from University of New South Wales (UNSW), Sydney, Australia, in 2019. She is currently working towards the Ph.D. degree in the School of Electrical Engineering and Telecommunications, UNSW, Sydney. Her research interests are image compression and machine learning.

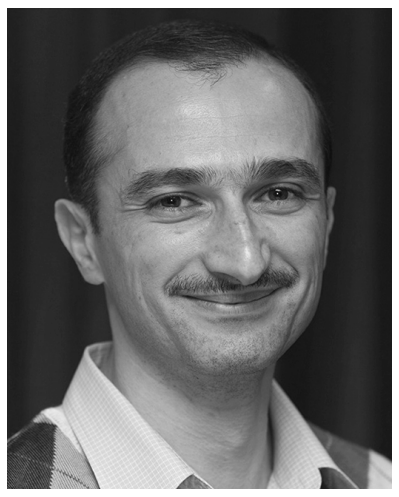

**Aous Naman** Aous Naman received the B.Sc. degree in electronics and telecommunication engineering from Al-Nahrain University, Baghdad, Iraq, in 1994, the M.Eng.Sc. degree in engineering from the University of Malaya, Kuala Lumpur, Malaysia, in 2000, and the Ph.D. degree in electrical engineering from the University of New South Wales (UNSW), Sydney, Australia, in 2011. Since then, he has been working as a post-doctoral researcher with the School of Electrical Engineering and Telecommunications, UNSW. He contributed to the development of the HTJ2K image coding standard, and he is also the main contributor to the OpenJPH open source implementation of HTJ2K. His research interests are in image or video compression and delivery.

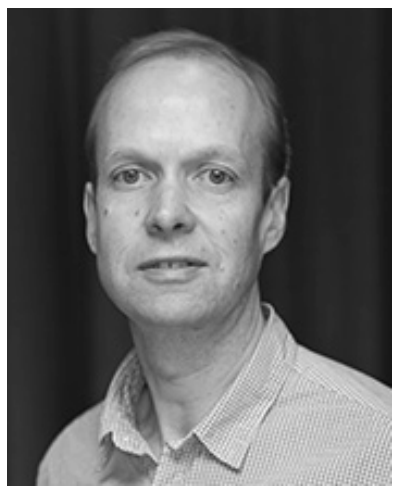

**David Taubman** David Taubman received the B.S. and B.E. (electrical) degrees from The University of Sydney, in 1986 and 1988, respectively, and the M.S. and Ph.D. degrees from the University of California at Berkeley, in 1992 and 1994, respectively. From 1994 to 1998, he was with Hewlett-Packard's Research Laboratories, Palo Alto, CA, USA, joining the University of New South Wales in 1998, where he is currently a Professor with the School of Electrical Engineering and Telecommunications. He has authored, along with M. Marcellin, the book, *JPEG2000: Image Compression Fundamentals, Standards and Practice.* His research interests include highly scalable image and video compression, motion estimation and modeling, inverse problems in imaging, perceptual modeling, and multimedia distribution systems. He received the University Medal from The University of Sydney. He has received two Best Paper Awards: from the IEEE Circuits and Systems Society for the 1996 paper, "A Common Framework for Rate and Distortion Based Scaling of Highly Scalable Compressed Video;" and from the IEEE Signal Processing Society for the 2000 paper, "High Performance Scalable Image Compression with EBCOT."